\pdfoutput=1

\documentclass[11pt]{article}

\usepackage[]{acl}

\usepackage{times}
\usepackage{latexsym}

\usepackage[T1]{fontenc}
\usepackage{booktabs} 
\usepackage{makecell} 
\usepackage{array} 
\usepackage{tabularx} 
\usepackage{graphicx}
\usepackage{adjustbox}
\usepackage{enumitem}
\usepackage{forest}
\usepackage{multirow}
\usepackage{multicol}
\usepackage{xspace}
\usepackage{amssymb}
\usepackage{subcaption}
\newcommand{\myquad}{\hspace{0.5em}}

\usepackage[utf8]{inputenc}

\usepackage{microtype}

\usepackage{inconsolata}

\usepackage{graphicx}

\title{Understanding Figurative Meaning through Explainable Visual Entailment}

\newcommand\tab[1][0.7cm]{\hspace*{#1}}
\newcommand{\emldisplay}[2]{\texttt{\href{mailto:#1}{#2}}}
\newcommand{\eml}[1]{\emldisplay{#1}{#1}}
\newcommand{\affa}{{$^{1}$}}

\author{Arkadiy Saakyan$^{1}$ \tab Shreyas Kulkarni\affa \tab Tuhin Chakrabarty\affa \tab Smaranda Muresan\affa \\
  \affa Columbia University \\
  \eml{a.saakyan@cs.columbia.edu}
}

\begin{document}
\maketitle
\begin{abstract}
  Large Vision-Language Models (VLMs) have demonstrated strong capabilities in tasks requiring a fine-grained understanding of literal meaning in images and text, such as visual question-answering or visual entailment. However, there has been little exploration of the capabilities of these models when presented with images and captions containing \textit{figurative} meaning, such as metaphors or humor. To close this gap, we propose a new task framing the \textit{figurative meaning understanding} problem as an \textit{explainable visual entailment} task, where the model has to predict whether the image (premise) entails a caption (hypothesis) and justify the predicted label with a textual explanation. The figurative phenomena can be present in the image, in the caption, or both. Using a human-AI collaboration approach, we build the accompanying expert-verified dataset \underline{V-FLUTE}, containing 6,027 $\{$image, caption, label, explanation$\}$ instances spanning five diverse figurative phenomena: metaphors, similes, idioms, sarcasm, and humor. Through automatic evaluation, we find that VLMs struggle to generalize from literal to figurative meaning, particularly when it is present in images. Further, we identify common types of errors in VLM reasoning (hallucination and incomplete or unsound reasoning) across classes of models via human evaluation.\footnote{Code and data: \href{https://github.com/asaakyan/V-FLUTE}{github.com/asaakyan/V-FLUTE}}

\end{abstract}

\section{Introduction}

\begin{figure}[!ht]
\centering
    \includegraphics[width=\columnwidth]{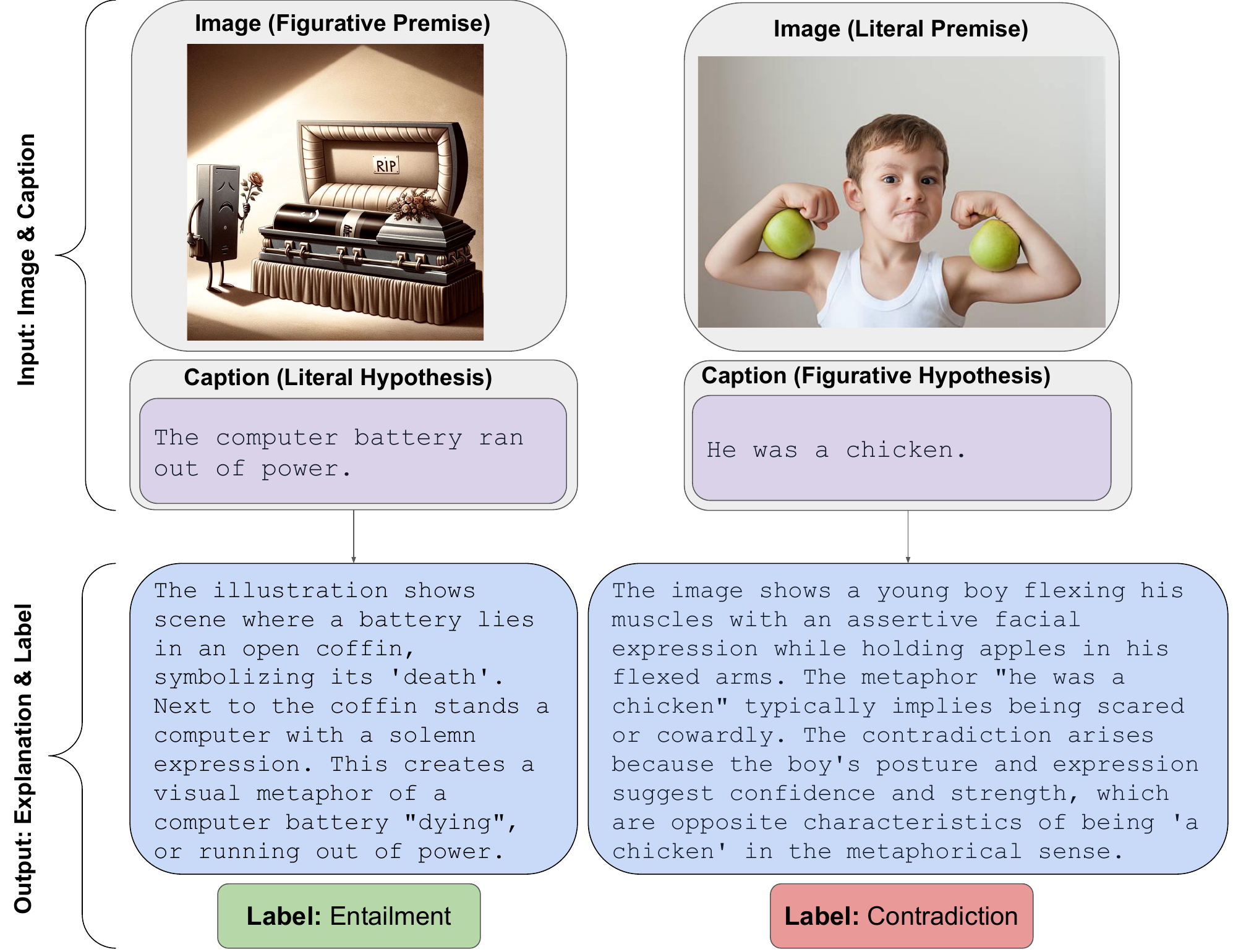}
    \caption{
    Explainable visual entailment for understanding figurative meaning: given an image and a caption output whether the image entails or contradicts the caption along with a textual explanation.
    }
    \label{fig:introFig}
\end{figure}

Figurative language is integral to human communication, enabling a variety of communicative goals \cite{doi:10.1111/j.1467-9280.1994.tb00653.x}, including affective communication \cite{fussell2014figurative}. Figurative language presents a significant challenge to computational approaches as it requires understanding of implicit meaning behind an expression \cite{stowe-etal-2022-impli, shutova2011computational, veale2016metaphor, zhou-etal-2021-pie}. 
Recently, \citet{chakrabarty-etal-2022-flute} proposed a task and dataset for Figurative Language Understanding through Textual Explanations (FLUTE) that frames the problem as an explainable textual entailment covering a variety of figurative language phenomena in text: metaphors, similes, idioms, and sarcasm. This dataset has been used successfully to advance and benchmark the capabilities of LLMs for understanding figurative language in text \cite{saakyan2022report,ziems2024can,sravanthi2024pub, dey2024socialite}. 

However, figurative meaning is also prevalent in visual phenomena, such as visual metaphors \cite{akula2023metaclue, chakrabarty-etal-2023-spy}, multimodal sarcasm \cite{desai2022nice}, and humor \cite{hessel-etal-2023-androids, hwang-shwartz-2023-memecap}. Yet so far most of the work on vision and language models (VLMs) has focused on understanding literal meaning in images and captions (e.g., ScienceQA \cite{lu2022learn}, MMMU \cite{yue2023mmmu}) including work on explainable visual entailment \cite{kayser2021vil}. Building on the idea of FLUTE \cite{chakrabarty-etal-2022-flute} for text, we present a new dataset for understanding figurative meaning as explainable visual entailment, V-FLUTE. Our dataset contains 6,027 $\{$image, caption, label, explanation$\}$ instances spanning diverse figurative phenomena.
Each instance contains an image (premise) and a caption (hypothesis) that is either entailed or contradicted by the image. Deciding the entailment relation requires the vision-language model to understand the implicit meaning in both the visual and textual modalities. Our dataset contains figurative phenomena present in the image, in the caption, or in both. In addition, to mitigate the dependence on spurious correlations, to more rigorously investigate reasoning capabilities, and to promote explainability, our task requires the model to generate a plausible explanation for the output label. See Figure \ref{fig:introFig} for two examples from our dataset. 

We make the following contributions towards assessing VLMs ability to understand figurative meaning expressed multimodally:
\begin{itemize}[leftmargin=*]
    \itemsep0em
    \item V-FLUTE, an expert-verified dataset of 6,027 $\{$image, caption, label, explanation$\}$ instances 
    built using a human-LLM collaboration framework covering several phenomena: metaphors, similes, idioms, sarcasm, and humor (Section \ref{taskdata}). 
    \item A suite of evaluations to assess current VLMs' capabilities on this new task of explainable visual figurative entailment (Section \ref{subsec:autoMetrics} and \ref{subsec:autoMetricsres}).
    \item A detailed human evaluation with error analysis yielding insights into the types of errors for different classes of models (Section \ref{sec:humanEval}).
\end{itemize}

\section{Related Work}
\begin{table*}[!ht]
\small
\centering
\begin{tabular}{@{}lllllr@{}}
\toprule
\textbf{Phenomenon} &
  \textbf{Data Source} &
  \textbf{Visual Style} &
  \textbf{Figurative Part} &
  \textbf{Our Contribution} &
  \multicolumn{1}{c}{\textbf{\# instances}} \\ \midrule
\multirow{2}{*}{\makecell{\textbf{Metaphor/} \\ \textbf{Simile}}} &
  \begin{tabular}[c]{@{}l@{}}HAIVMet\\ \cite{chakrabarty-etal-2023-spy}\end{tabular} &
  Illustration &
  Image &
  \begin{tabular}[c]{@{}l@{}}Image Selection \\
 Textual Explanations \\ Expert Verification \end{tabular} &
  \makecell{
  857  \\
  (450 E, 407 C)} \\ \cmidrule(l){2-6} 
 &
  \begin{tabular}[c]{@{}l@{}}IRFL\\ \cite{yosef-etal-2023-irfl}\end{tabular} &
  Photographic &
  Caption &
  \begin{tabular}[c]{@{}l@{}}Image Selection\\ Textual Explanations \\ Expert Verification \end{tabular} &
  \makecell{
  1,149  \\
  (574 E, 575 C)} \\ \midrule
\textbf{Idiom} &
  \begin{tabular}[c]{@{}l@{}}IRFL\\ \cite{yosef-etal-2023-irfl}\end{tabular} &
  Photographic &
  Caption &
  \begin{tabular}[c]{@{}l@{}}Image Selection\\ Textual Explanations \\  Expert Verification \end{tabular} &
  \makecell{
  370  \\
  (186 E, 184 C)} \\ \midrule
\textbf{Sarcasm} &
  \begin{tabular}[c]{@{}l@{}}MuSE\\ \cite{desai2022nice}\end{tabular} &
  Meme &
  Caption &
  \begin{tabular}[c]{@{}l@{}}Caption Generation\\ Textual Explanations \\ Expert Verification \end{tabular} &
  \makecell{
  1,042  \\
  (521 E, 521 C)} \\ \midrule
\multirow{2}{*}{\textbf{Humor}} &
  \begin{tabular}[c]{@{}l@{}}MemeCap\\ \cite{hwang-shwartz-2023-memecap}\end{tabular} &
  Meme &
  Image &
  \begin{tabular}[c]{@{}l@{}}Caption Generation\\ Textual Explanations \\ Expert Verification \end{tabular} &
  \makecell{
  1,958  \\
  (979 E, 979 C)} \\ \cmidrule(l){2-6} 
 &
  \begin{tabular}[c]{@{}l@{}}NYCartoons\\ \cite{hessel-etal-2023-androids}\end{tabular} &
  Illustration &
  Image+Caption &
  Taken As Is &
  \makecell{
  651  \\
  (651 E)} \\ \bottomrule
\end{tabular}
\caption{V-FLUTE dataset composition: 5 figurative phenomena, source datasets, visual styles, and our contributions. E denotes number of entailment instances, C - contradiction. Diversity of the dataset ensures coverage of various figurative phenomena, figurative meaning location, and visual styles.}
\label{tab:datasets}
\end{table*}

Textual entailment \cite{maccartney-manning-2008-modeling, bowman-etal-2015-large} and visual entailment \cite{Xie2019VisualEA} tasks have been proposed to measure language and multimodal understanding.
However, models trained to simply improve label accuracy on these data can be brittle and suffer from spurious correlations \cite{poliak-etal-2018-hypothesis, gururangan-etal-2018-annotation, mccoy-etal-2019-right, gardner-etal-2021-competency}.
Datasets such as e-SNLI \cite{eSNLI} and e-SNLI-VE \cite{kayser2021vil} augment existing entailment datasets with natural language explanations and train models to not only predict the label, but also generate a textual explanation for the reason behind the prediction. However, they only focus on \emph{literal meaning} in text and images.
Recently, explainable entailment has been utilized to assess LLMs' capabilities on understanding figurative language through the FLUTE dataset \cite{chakrabarty-etal-2022-flute}. FLUTE frames figurative language understanding as an explainable textual entailment task.
Recent progress in multimodal models \cite{ li2022blip, alayrac2022flamingo, gpt4v, team2023gemini, visInstrTune, claude3} prompts us to asses understanding of figurative meaning present in the multimodal setting, contained in both images and text beyond intent and sentiment \cite{zhang-etal-2021-multimet, kruk-etal-2019-integrating}. 
To this end, we present an equivalent of the FLUTE dataset for the visual modality: V-FLUTE.


\begin{table*}[ht]
\small
\centering
\setlength{\tabcolsep}{1pt}
\begin{tabular}{ccccc}
\hline
HAIVMet & IRFL & MuSE & MemeCap & NYCartoons \\ \hline
\raisebox{-\totalheight}{\includegraphics[scale=0.08]{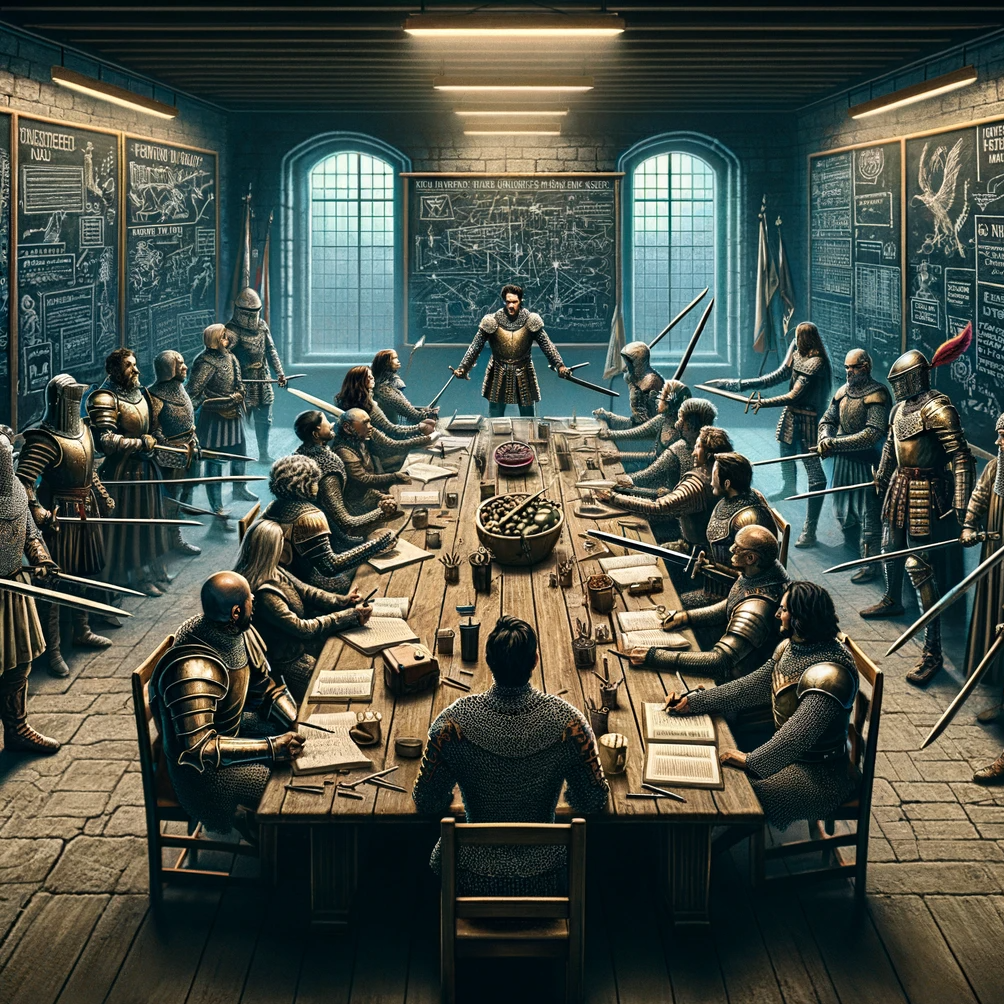}}   & \raisebox{-\totalheight}{\includegraphics[scale=0.30]{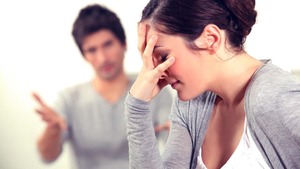}}   & \raisebox{-\totalheight}{\includegraphics[scale=0.25]{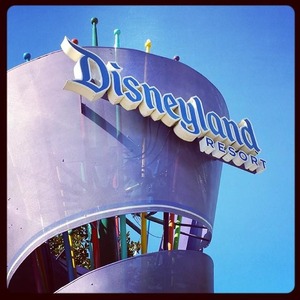}}   & \raisebox{-\totalheight}{\includegraphics[scale=0.30]{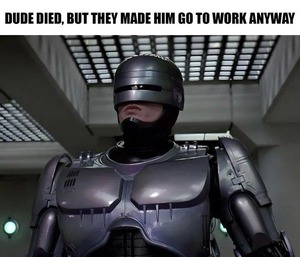}}   & \raisebox{-\totalheight}{\includegraphics[scale=0.28]{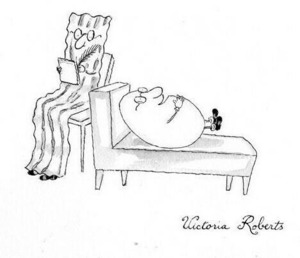}}   \\\\ \hline
\begin{tabular}[c]{@{}l@{}}The faculty meeting \\was peaceful.\end{tabular} & \begin{tabular}[c]{@{}l@{}}Their relationship is\\ a house on fire.\end{tabular} & \begin{tabular}[c]{@{}l@{}}Oh I just \#love \\having to stare at\\ this while I \#work.\end{tabular} & \begin{tabular}[c]{@{}l@{}}Even death won't \\exempt you from \\going to work.\end{tabular} & \begin{tabular}[c]{@{}l@{}}Easy for you to\\ say, you're cured!\end{tabular} \\ \hline
Contradiction & Entailment & Contradiction & Entailment & Entailment \\ \hline
\begin{tabular}[c]{@{}l@{}}The image shows \\a faculty meeting \\ transformed into a \\dramatic battlefield ...\\ The visual metaphor \\suggests the faculty \\meeting was like a \\war, and not peaceful.\end{tabular} & \begin{tabular}[c]{@{}l@{}}The photo suggests \\a conflict or an \\intense emotional \\situation ... which aligns \\ with the symbolism of a \\ house on fire representing \\ a relationship filled \\ with turmoil or \\ heated arguments.\end{tabular} & \begin{tabular}[c]{@{}l@{}}
The image shows\\ Disneyland\\ Resort sign ... the \\person would like\\ to experience it \\in person rather\\ than just looking\\ at the sign during\\ work hours.
\end{tabular} & \begin{tabular}[c]{@{}l@{}}
The image shows \\RoboCop ...\\ it humorously\\ illustrates a \\character who has\\ been reanimated\\ as a cyborg to\\ continue working \\despite having died.
\end{tabular}
 & \begin{tabular}[c]{@{}l@{}}
A play on the word \\ "cured". People seek \\ therapy to have their \\ mental problems \\remedied or cured. \\But "cured" can also\\refer to a meat prep\\ technique ...
\end{tabular}
 \\ \hline
\end{tabular}
\caption{Sample dataset instances form V-FLUTE corresponding to the source datasets displaying images (premise), captions (hypothesis), labels, and explanations [Row 1-5]. }
\label{table:sample_data}
\end{table*}

\section{V-FLUTE Task and Dataset}\label{taskdata} 

Following prior work on figurative language understanding in text defined as explainable textual entailment, FLUTE \cite{chakrabarty-etal-2022-flute}, we define  \emph{understanding figurative meaning} as an \emph{explainable visual entailment task}: given an image (premise) $p$ and a caption (hypothesis) $h$, output a textual explanation $\hat{e}$ justifying whether the premise entails or contradicts the hypothesis and assign a label $\hat{y} \in \{ \texttt{Entailment}, \texttt{Contradiction} \}$. We focus on the binary classification task, since for neutral labels, the explanations would be trivial (simply describing the image).

To build V-FLUTE, we start with existing multimodal figurative datasets 
which cover phenomena such as metaphors, similes, idioms, sarcasm or humor. We utilize human-AI collaboration frameworks with expert annotators \cite{chakrabarty-etal-2022-flute, wiegreffe-etal-2022-reframing, liu-etal-2022-wanli} to augment them with expert-verified textual explanations and entailing/contradicting captions. Each instance then includes an image and a caption, and the figurative phenomenon can be either in the image, the caption or in both. 
An overview of the V-FLUTE dataset and \emph{our contributions w.r.t to the source datasets can be found in Table \ref{tab:datasets}.} See examples corresponding to each source dataset in Table \ref{table:sample_data} as they appear in V-FLUTE. Below, we describe the construction of V-FLUTE by each phenomenon.

\subsection{Metaphors, Similes and Idioms}
To create visual entailment instances containing metaphors and similes in V-FLUTE, we rely on two existing resources: HAIVMet \cite{chakrabarty-etal-2023-spy} and IRFL \cite{yosef-etal-2023-irfl}. Instances from HAIVMet contain the metaphor/simile as a part of the premise (image), while those taken from IRFL have the metaphor/simile as a part of the hypothesis (text). 

\subsubsection{IRFL as Data Source} \label{sec:irfl}

\begin{figure}[!ht]
\centering
    \includegraphics[width=0.95\columnwidth]{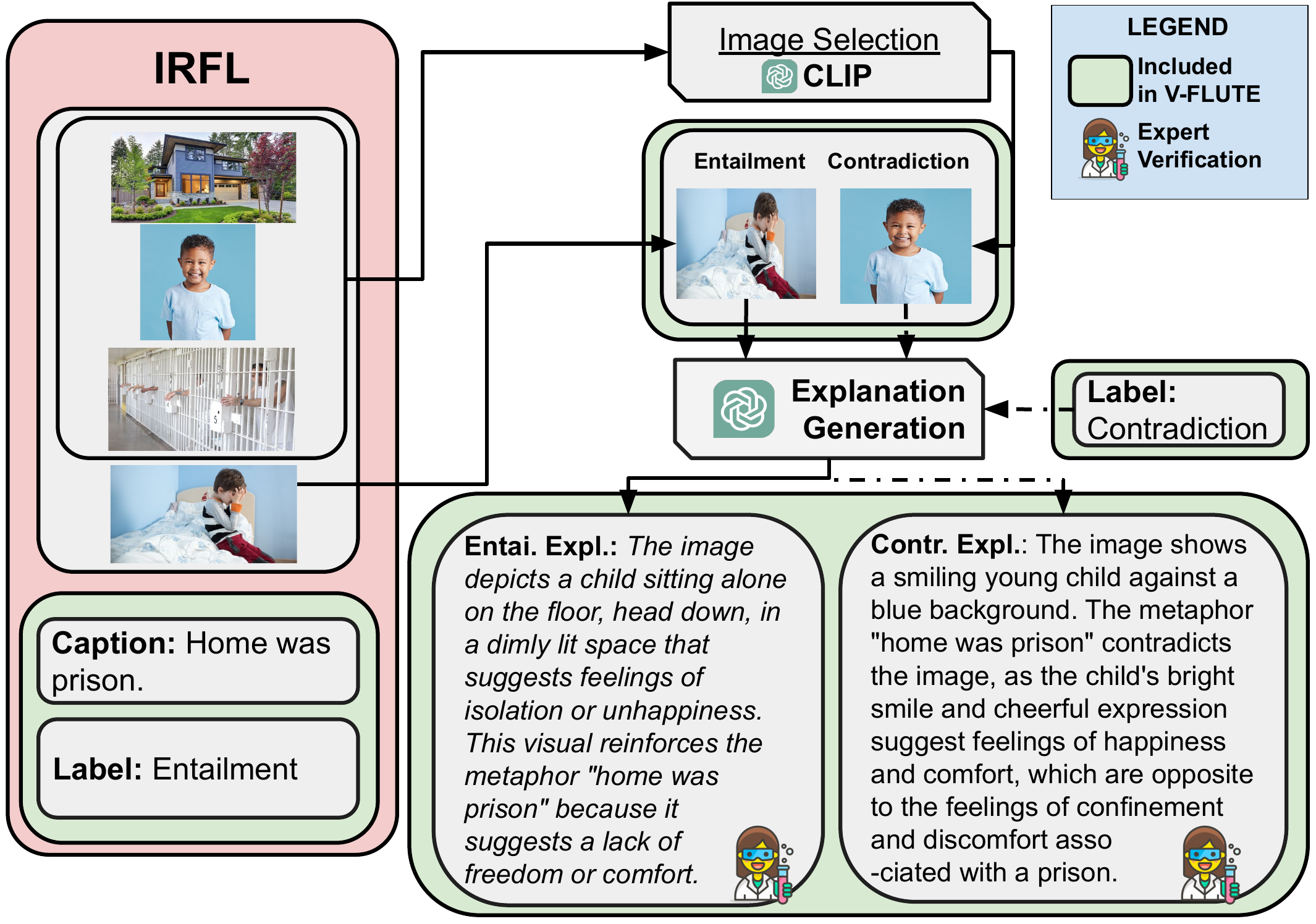}
    \caption{Creation of V-FLUTE instances for metaphors, similes, idioms from IRFL.}
    \label{fig:irfl}
\end{figure}


\citet{yosef-etal-2023-irfl} proposed a benchmark (IRFL) where given a metaphor, a simile or an idiom the model has to distinguish which of the four associated images implies the figurative meaning of the expression. This dataset contains 1,440 figurative expressions, each associated with $4$ distinct images. One of those images represents the figurative expression (see Figure \ref{fig:irfl}), and the other 3 act as distractors. 

{\bf Image Selection.} We automatically select images using CLIP \cite{radford2021learning}. We select one of the distractor images that have the highest CLIPScore (\texttt{clip-vit-base-patch16}) with the corresponding entailing image to create a challenging, contradictory instance (see where an unrelated image of a house is discarded when selecting the contradiction instance in Figure \ref{fig:irfl}). 

{\bf Generating Textual Explanations.} We prompt GPT-4 (\texttt{gpt-4-vision-preview}) with the ground truth label, caption, and the image to explain the relationship between the image and the caption.

{\bf Expert Verification.} 
We recruit three expert annotators with significant experience in figurative language and visual metaphor understanding on Upwork and ask them to verify the explanation is correct, complete, and concise and if not, edit it (see details in Appendix \ref{app:annotDetails}). We also ask the annotators to discard rare noisy instances where the caption, image, and label do not fit (due to automatic image selection). Due to relative simplicity of generating the explanation given a literal image, the experts only needed to edit $\approx 7$\% of the explanations. They also removed $\approx 1\%$ the data, resulting in 1149 $\{$image, caption, label, explanation$\}$ instances for metaphors and similes and 370 for idioms. 


\subsubsection{HAIVMet as Data Source}


\begin{figure}[!ht]
\centering
    \includegraphics[width=0.9\columnwidth]{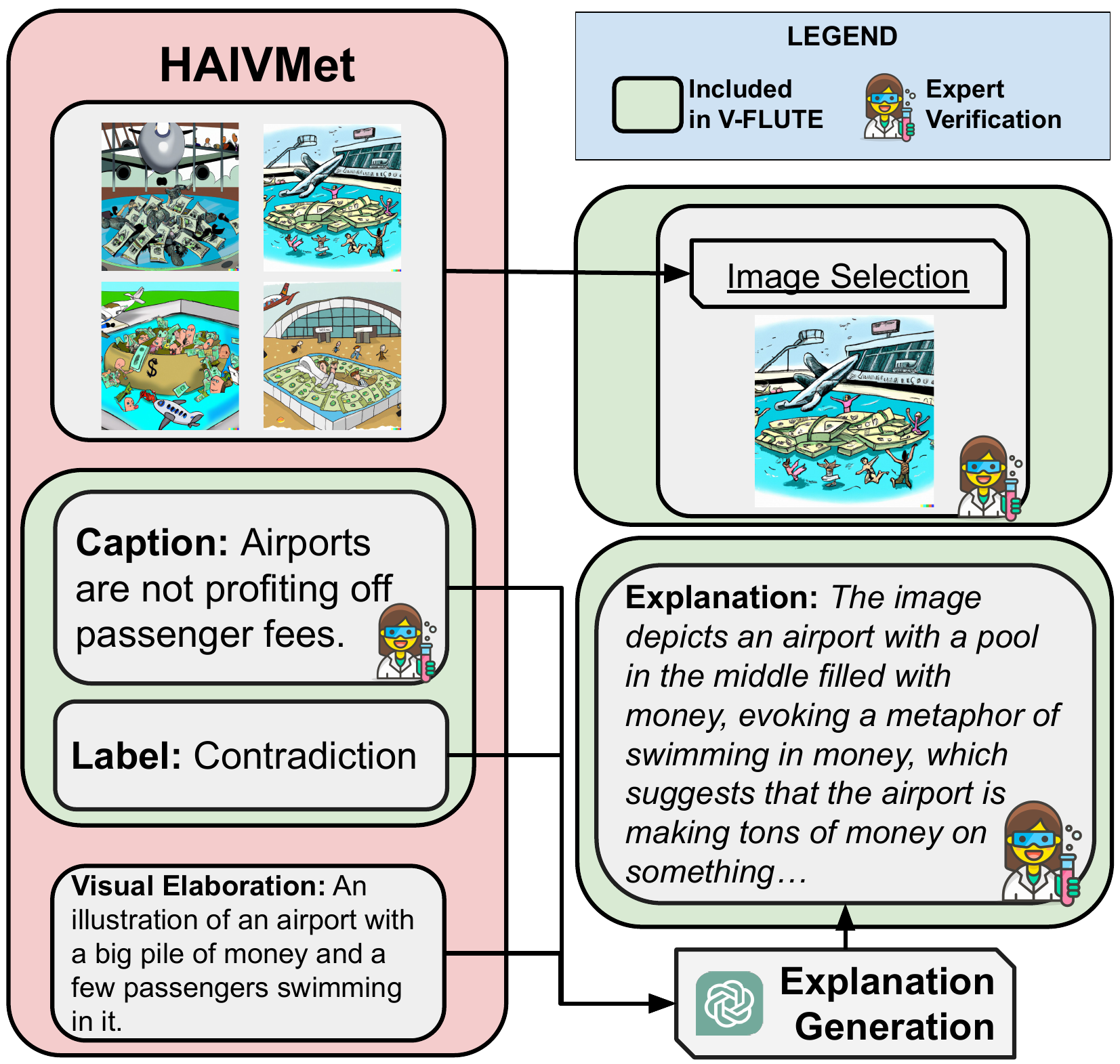}
    \caption{Creation of V-FLUTE instances for metaphors and similes from HAIVMet.}
    \label{fig:haivmet}
\end{figure}

\citet{chakrabarty-etal-2023-spy} use a human-AI collaboration framework to generate visual metaphors from linguistic metaphors (HAIVMet dataset) and propose a visual entailment task as an extrinsic evaluation of dataset quality. The HAIVMet data consists of 1,193 images of visual metaphors spanning over 958 distinct linguistic metaphors. Each image is associated with a caption that can be contradicting or entailing the image.
In addition, each image is associate with
a \textit{visual elaboration} that presents a textual description of the image (See Figure \ref{fig:haivmet}). This visual elaboration was used in the original paper to generate the visual metaphors (images). 

{\bf Generating Textual Explanations.} We augment the dataset with candidate textual explanations.
We prompt ChatGPT (\texttt{gpt-3.5-0914}) to generate an explanation for every tuple $\{$visual elaboration, caption, label$\}$
(See Figure \ref{fig:haivmet}; and prompt in Appendix \ref{subsubsec:vismet-gen-explain}). 

{\bf Expert Verification.} Each caption is paired with up to $5$ images. However, since these images were automatically generated with DALLE-2 using the visual elaborations, not all are completely faithful. Moreover, some captions and labels were inconsistent. Finally, automatically generated LLM candidate explanations are not always correct and require refining. To tackle these issues, we employ an expert verification process recruiting the same three expert annotators as from the IRFL section above (see details in Appendix \ref{app:annotDetails}). We ask the annotators to select the visual metaphor most faithful to the linguistic metaphor and the visual elaboration (see \underline{Image Selection} in Figure \ref{fig:haivmet}) or if none were. In addition, we ask them to verify and edit the explanation if necessary to ensure correctness, completeness, and conciseness. On average, experts edited $\approx 65\%$ of the explanations and $29 \%$ of captions, and rejected $\approx 30\%$ of visual metaphors, resulting in 857 $\{$image, caption, label, explanation$\}$ instances. 

\subsection{Sarcasm}

To create visual entailment instances containing sarcasm, we rely on the MuSE data \cite{desai2022nice}.  
\subsubsection{MuSE as Data Source} 
\begin{figure}[htbp]
    \centering
    \includegraphics[width=0.9\columnwidth]{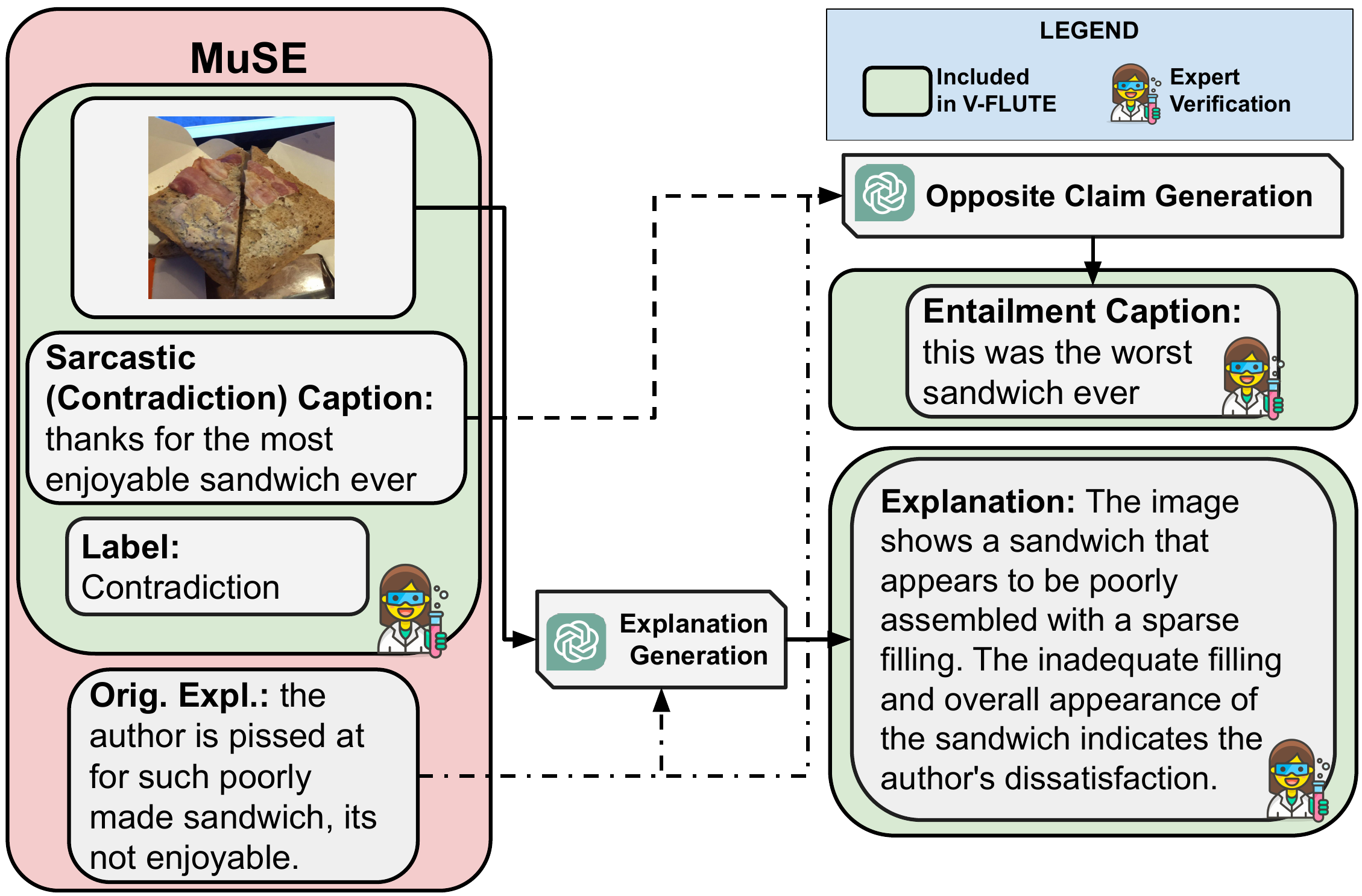}
    \caption{Creation of V-FLUTE instances for sarcasm from MuSE.}
    \label{fig:museFig}
\end{figure}
The MuSE dataset \cite{desai2022nice} consists of 3510 distinct images, the respective sarcastic captions that act as contradiction instances (see example in Figure \ref{fig:museFig}), and crowd worker written explanations justifying the contradiction. 

{\bf Generating Entailment Captions.}
Since the dataset only contains sarcastic instances, there are no captions with an entailment relationship. We generate the entailing captions by prompting GPT-4 to generate a non-sarcastic version of the caption while maintaining the user-generated informal style of the text (see the generated entailment caption in Figure \ref{fig:museFig}). 

{\bf Generating Textual Explanations.} While the dataset already contains crowdworker-written explanations, upon inspection, they were often deemed poor quality, lacking enough details, and formulaic (e.g., see the crowdworker explanation in Figure \ref{fig:museFig}). To improve their quality, we use the dataset's existing crowdworker explanations and prompt GPT-4 to rewrite and generate candidate textual explanations given the caption and the label (see the re-written explanation in Figure \ref{fig:museFig}). See the prompt in Appendix \ref{sub:muse-prompts}.

{\bf Expert Verification.} Each image is now paired with a GPT-4-generated entailing caption, an original contradicting caption, and their respective labels and explanations. 
The same three expert annotators checked if the generated explanations are adequate (i.e., complete, correct, and concise) and if not, asked to edit them. The experts were also instructed to discard noisy examples, e.g. when the image does not contradict the sarcastic caption. On average, experts edited $\approx 13\%$ of the initial explanations and rejected $\approx 18\%$ of the examples, resulting in 
1,042 $\{$image, caption, label, explanation$\}$ instances.  

\subsection{Humor}
 For multimodal humor, we rely on two datasets: MemeCap \cite{hwang-shwartz-2023-memecap} and New Yorker cartoons \cite{hessel-etal-2023-androids}.

\subsubsection{MemeCap as Data Source} 

\begin{figure}[htbp]
    \centering
    \includegraphics[width=0.9\columnwidth]{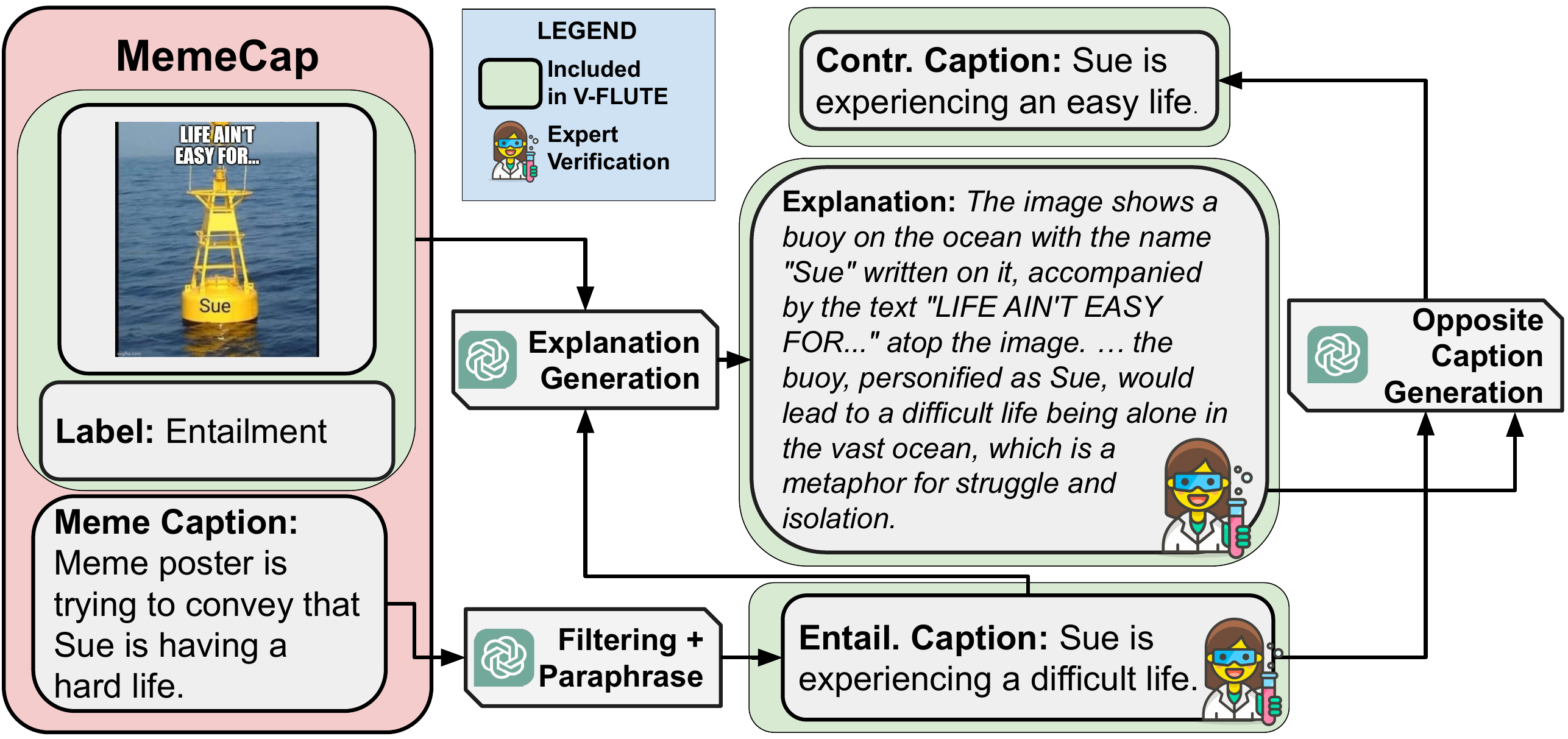}
    \caption{Creation of V-FLUTE instances for humor from MemeCap.}
    \label{fig:memecapFig}
\end{figure}

This dataset consists of memes along with their captions that describe the meme poster's intent (see example in Figure \ref{fig:memecapFig}). Memes frequently contain implicit, non-literal meaning \cite{pioneer} and rely on visual metaphors \cite{PIATA201639}, posing a challenge to VLMs.

{\bf Caption Generation.} Meme captions are not suited for an entailment task, so we prompt GPT-4 with the original caption to generate an entailing caption in the form of a claim from it (see example in Figure \ref{fig:memecapFig}). We filter these set of samples further with GPT-4 by asking whether the image entails the caption and only selecting positive instances. In addition to generating captions that entail the meme, we generate contradicting captions using GPT-4.

{\bf Generating Textual Explanations.} We prompted GPT-4 with the ground truth label in the prompt to explain the relationship between the image and the caption. See prompts in Appendix \ref{subsec:memecap-prompts}.

{\bf Expert Verification.} We hire the same three expert annotators to ensure the correctness of the data. Each annotator is tasked with verifying that 1) the generated caption fits the image and 2) the explanation is correct and complete, and if not, make the necessary changes. We also ask to discard samples with inappropriate content. Experts edited $\approx 35\%$ of the explanations and $15 \%$ of captions on average, and discarded $\approx 2\%$ of inappropriate instances, resulting in 1958 $\{$image, caption, label, explanation$\}$ instances.

\subsubsection{NYCartoons as Data Source} The NYCartoons dataset \cite{hessel-etal-2023-androids} contains 651 high-quality instances from the New Yorker Cartoon Caption Contest. Each instance consists of an image paired with a humorous caption and an explanation of why this combination of the caption and the image is funny. We utilize this data as is by treating the image as entailing the caption, so the explanation of the entailment relationship is the explanation of the joke.

\subsection{Dataset Statistics}
We split our data into 4,578 training, 726 validation, and 723 testing instances.  Table \ref{tab:dataCounts} shows the number of samples from each source dataset that are included in the randomly selected training, validation, and held-out test splits. More details in Appendix \ref{app:dataStats}.

\begin{table}[htbp]
\centering
\small
\begin{tabular}{@{}lllll@{}}
\toprule
\textbf{Type} & \textbf{Dataset} & \textbf{Train} & \textbf{Valid} & \textbf{Test} \\ \midrule
\multirow{2}{*}{\textbf{\begin{tabular}[c]{@{}l@{}}Metaphor\\ /Similes\end{tabular}}} & HAIVMET & 649 & 107 & 101 \\ \cmidrule(l){2-5} 
 & \begin{tabular}[c]{@{}l@{}}IRFL \\ (metaphor\\ /simile)\end{tabular} & 912 & 117 & 120 \\ \midrule
\textbf{Idioms} & IRFL (idiom) & 170 & 100 & 100 \\ \midrule
\textbf{Sarcasm} & MuSE & 830 & 106 & 106 \\ \midrule
\multirow{2}{*}{\textbf{Humor}} & MemeCap & 1566 & 196 & 196 \\ \cmidrule(l){2-5} 
 & NYCartoons & 451 & 100 & 100 \\ \midrule \midrule
\multicolumn{2}{c}{\textbf{Total}} & 4,578 & 726 & 723 \\ \bottomrule
\end{tabular}
\caption{Data counts per phenomenon and dataset.}
\label{tab:dataCounts}
\end{table}

\section{Experiments}

We empirically study how several baseline models perform on the task of explainable visual entailment. We investigate both off-the-shelf and fine-tuned model performance. We provide \emph{human baseline performance} in Appendix \ref{subsec:humanBaseline}. \emph{Hyperparameters} are provided in Appendix \ref{app:hyper}. 

\subsection{Models}

We select a variety of models for our study (see taxonomy in Appendix, Figure \ref{fig:modelTaxonomy}). For \textbf{off-the-shelf models}, we explore both \textit{open} and \textit{API-based} models. For \textit{open} models, we select the (current) state-of-the-art LLaVA-1.6 models \cite{liu2024llavanext}. LLaVA is one of the simplest, yet one of the most high-performing VLM architectures currently available. It utilizes a pretrained large language model (e.g., Mistral-7B \cite{jiang2023mistral}) and a vision-language cross-modal connector (e.g., an MLP layer) to align the vision encoder (e.g., CLIP \cite{radford2021learning}) outputs to the language models. We select LLaVA-1.6 models in their 7B and 34B configurations (LLaVA-v1.6-7B and LLaVA-v1.6-34B respectively) and refer to them as \textit{LLaVA-ZS-7B} and \textit{LLaVA-ZS-34B}. Both models have been instruction-tuned on less than 1M visual instruction tuning samples to act as general language and vision assistants. 
We also utilize \textit{Compositional Chain-of-Thought Prompting} proposed by \citet{mitra2023compositional} denoted by \underline{LLaVA-ZS-7B-SG} and \underline{LLaVA-ZS-34B-SG} (see description and results discussion in Appendix \ref{app:CoT}).

For \textit{API-based} models, we select three widely available state-of-the-art VLMs: Claude-3 Opus (\texttt{claude-3-opus-20240229})\cite{claude3}, GPT-4 (\texttt{gpt-4-1106-vision-preview}) \cite{gpt4v} and GeminiPro (\texttt{gemini-pro-vision})\cite{team2023gemini}. 

For \textbf{fine-tuned} models, we focus on fine-tuning the LLaVA-1.5-7B model\footnote{Fine-tuning code for 1.6 model was not published as of writing of this paper.} \cite{liu2023improvedllava}. To minimize bias for a single instruction, we fine-tune and evaluate the models on a set of 21 instruction paraphrases (see Appendix Table \ref{tab:instructs}). Three model configurations are tested:
\begin{itemize}[leftmargin=*]
    \itemsep0em
    \item \textit{LLaVA-VF} is the same checkpoint fine-tuned on the training set of V-FLUTE. We also fine-tune the model with a white square instead of the V-FLUTE image (denoted by $-$Image).
    \item \textit{LLaVA-eViL} and \textit{LLaVA-eViL+VF} are checkpoints of LLaVA-v1.5-7B further fine-tuned on the eViL (e-SNLI-VE) dataset for explainable visual entailment \cite{kayser2021vil} converted to the instruction format or on both eViL and V-FLUTE. We removed neutral label instances, which resulted in 275,815 training instances and 10,897 validation instances. 
    
\end{itemize} 

\subsection{Automatic Metrics} \label{subsec:autoMetrics}
 Since our goal is to ensure models provide an answer for the right reasons, ideally, we would only count predictions as correct when the explanation is also correct. Based on prior work \cite{chakrabarty-etal-2022-flute}, we use both the standard F1 score and an adjusted score that accounts for explanation quality: F1@ExplanationScore. The ExplanationScore computes the average of BERTScore \cite{bertscore} 
 and BLEURT \cite{sellam-etal-2020-bleurt} 
 between model-generated and reference (V-FLUTE) explanations.
 We report F1@0 (simply F1 score), F1@53\footnote{Thresholds selected based on human evaluation of explanation quality in Section \ref{subsec:explScoreCorr}.} (all predictions with ExplanationScore $\leq$ 53 are considered incorrect) and F1@60.

\subsection{Automatic Evaluation Results} \label{subsec:autoMetricsres}
We include results \emph{per phenomenon} 
in Appendix \ref{app:byPhen}, discussion on \emph{CoT prompting} in Appendix \ref{app:CoT} and \emph{additional models} in Appendix  \ref{app:addlModels}. Table \ref{tab:mainRes} shows the results, informing the following insights: 

\begin{table}[!t]
\small
    \centering
        \begin{tabular}{lccc}
        \toprule
        Model Name & F1@0 & F1@53 & F1@60 \\
        \midrule
        \textit{Random Baseline} & 49.82 & - & - \\
        \midrule
        \midrule
        \multicolumn{4}{l}{\textit{Fine-tuned}} \\
        LLaVA-7B & & & \\
        $\dashrightarrow$ VF & 72.78 & 60.66 & 47.12 \\
        \myquad $\dashrightarrow$ $-$ Image & 64.77 & 53.28 & 39.37 \\
        $\dashrightarrow$ eViL & 54.34 & 4.11 & 0.55 \\
        \myquad $\dashrightarrow$ $+$ VF & \underline{\textbf{74.91}} & \underline{\textbf{62.34}} & \underline{48.80} \\
        \midrule
        \midrule
        \multicolumn{4}{l}{\textit{Off-the-shelf}} \\
        \multicolumn{4}{l}{\underline{\textit{Open}}} \\
        LLaVA-ZS & & & \\
        $\dashrightarrow$ 7B & 45.44 & 35.57 & 18.38 \\
        \myquad $\dashrightarrow$ $+$ SG & 52.94 & 39.27 & 14.86 \\
        $\dashrightarrow$ 34B & 55.60 & \underline{48.32} & \underline{31.83} \\
        \myquad $\dashrightarrow$ $+$ SG & \underline{58.08} & 45.74 & 26.77 \\
        \midrule
        \multicolumn{4}{l}{\underline{\textit{API-based}}} \\
        Gemini-1.5-Pro & 53.70 & 39.72 & 19.01 \\
        \myquad $\dashrightarrow$ 5-shot & 67.25 & 56.04 & 37.14  \\
        Claude-3 Opus & 56.07 & 45.37 & 22.31 \\
        \myquad $\dashrightarrow$ 5-shot & 67.79 & 58.70 & 35.32 \\
        GPT-4 & 64.00 & 56.22 & 38.56 \\
        \myquad $\dashrightarrow$ 5-shot & \underline{69.36} & \underline{61.95} & \underline{\textbf{49.81}} \\
        \bottomrule
        \end{tabular}
    
    \caption{F1 Score results for different models across thresholds 0.0, 0.53, and 0.6 for explanation score. Best result overall is in bold, best result in each category is underlined.}
    \label{tab:mainRes}
\end{table}

\paragraph{A literal visual entailment dataset does not solve the figurative visual entailment task.}
Fine-tuning only on e-ViL barely improves over a random baseline (54.34 F1@0) and underperforms compared with the models fine-tuned on V-FLUTE (72.78 F1@0). Moreover, the explanations are of poor quality (0.55 F1@60). \emph{This indicates that models trained on a literal visual entailment task struggle to generalize to figurative meaning, supporting the challenging nature of our dataset.}

The strongest model fine-tuned on V-FLUTE (LLaVA-7B-eViL+VF) outperforms the best off-the-shelf model (GPT-4-5shot) in terms of the F1@0 score ($p<0.03$\footnote{$p$ values reported via paired bootstrap test \cite{koehn-2004-statistical}}). It performs competitively when incorporating the reference-based ExplanationScore, with GPT-4 leading slightly as it is the model with which the candidate explanations were generated. 

\begin{figure}[!ht]
\centering
    \includegraphics[width=0.9\columnwidth]{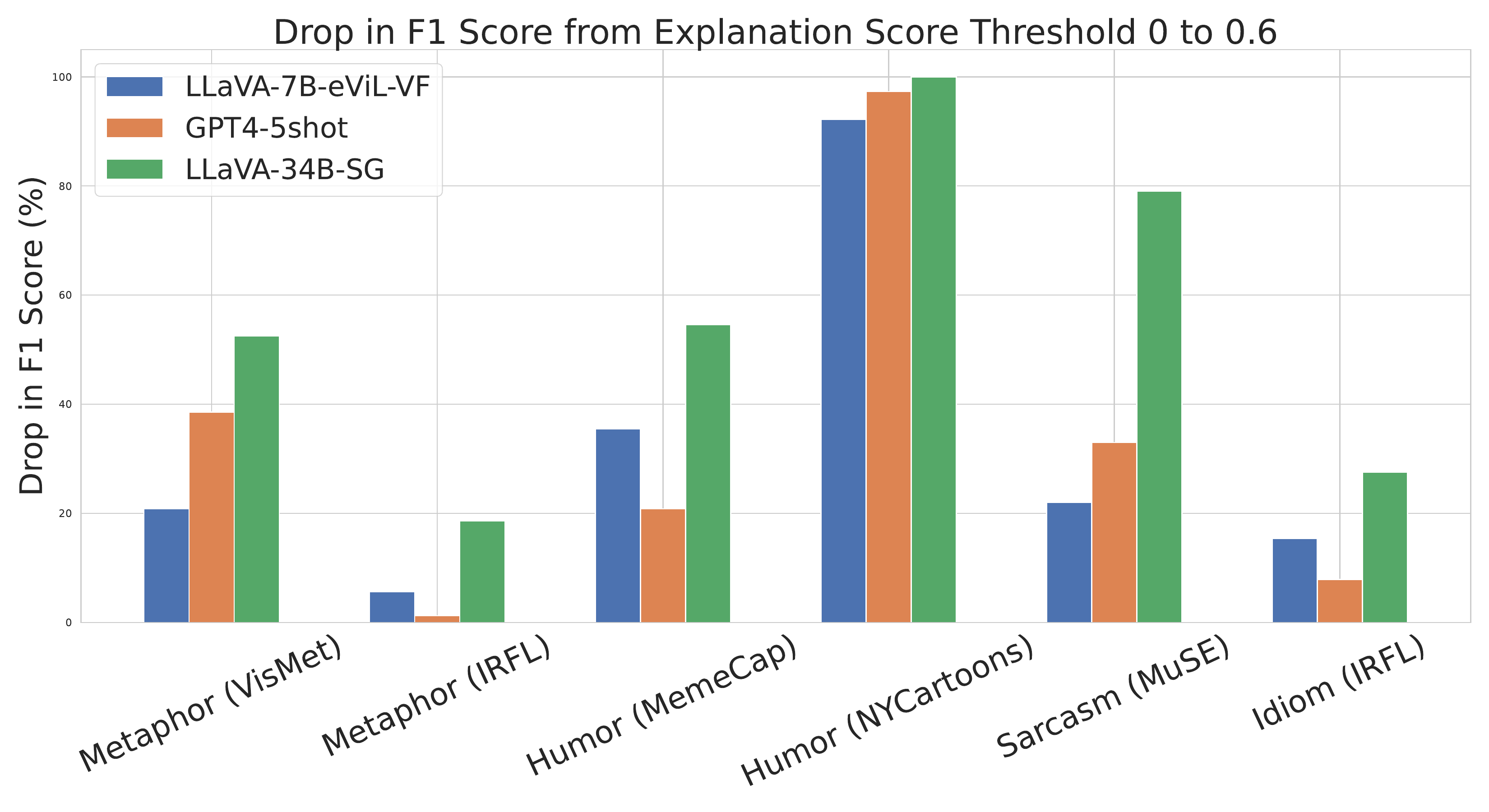}
    \caption{\% Drop in F1 score for various models by source dataset between 0 to 0.6. Higher drop indicates higher proportion of wrongly generated explanations.}
    \label{fig:f1Drop}
\end{figure}

\paragraph{When figurative meaning is in the image rather than text, models perform worse.} We plot the relative percentage decrease between F1@0 and F1@60 for LLaVA-eViL-VF, LLaVA-34B-SG, and GPT-4-5shot in Figure \ref{fig:f1Drop}. Higher performance drop indicates higher difficulty of generating the correct explanation. For all models, we see a substantial decrease in performance, especially on challenging phenomena such as Humor (NYCartoons). 
The percentage drop is substantially higher for all models for the HAIVMet subset rather than the IRFL dataset, which contains metaphors in the image rather than in the text. \textit{This suggests it is harder for models to generate correct explanations when the figurative meaning is contained in the image rather than in the text, indicating the need to expand the presence of figurative phenomena in existing visual datasets.}

\paragraph{VLMs benefit from visual information when dealing with figurative phenomena and do not just rely on the input text to make their prediction.} We utilize a hypothesis-only baseline \cite{poliak-etal-2018-hypothesis} by including a model fine-tuned on the V-FLUTE dataset, but with a white square as the image input, denoted as $-$Image. Fine-tuning on the full V-FLUTE dataset shows an improvement of over 8 points in F1@0 (better with $p<0.002$).

\section{Human Evaluation and Error Analysis} \label{sec:humanEval}

\begin{table*}[htbp]
\small
\centering
\begin{adjustbox}{width=\textwidth,totalheight=\textheight,keepaspectratio}
\begin{tabularx}{\textwidth}{p{2.5cm}>{\centering\arraybackslash}p{3.5cm}p{2.5cm}X}
\toprule
\textbf{Error Type} & \textbf{Image} & \textbf{Caption} & \textbf{Label and Explanation} \\
\midrule
\textbf{Hallucination} \textit{(describes sharp pencil as having a blunt tip)} & 
\raisebox{-\totalheight}{\includegraphics[scale=0.06]{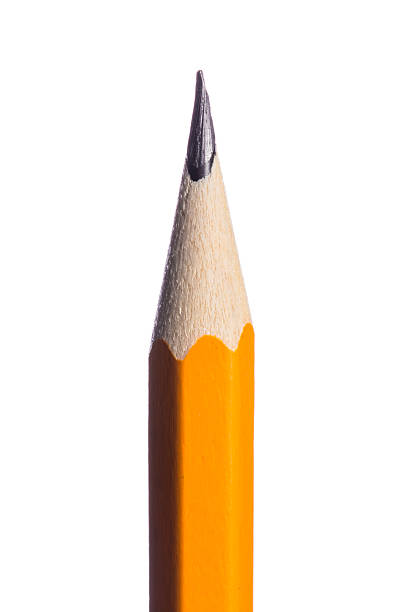}} & 
The tip is as sharp as a knife. & 
\textbf{Predicted Label:} \textcolor{red}{Contradiction} \newline 
\textit{Predicted Explanation:} The image depicts a pencil with a \textcolor{red}{blunt tip}. [...] \\
\midrule
\textbf{Incomplete} \textit{(does not address metaphorical meaning of iceberg imagery)} & 
\raisebox{-\totalheight}{\includegraphics[scale=0.06]{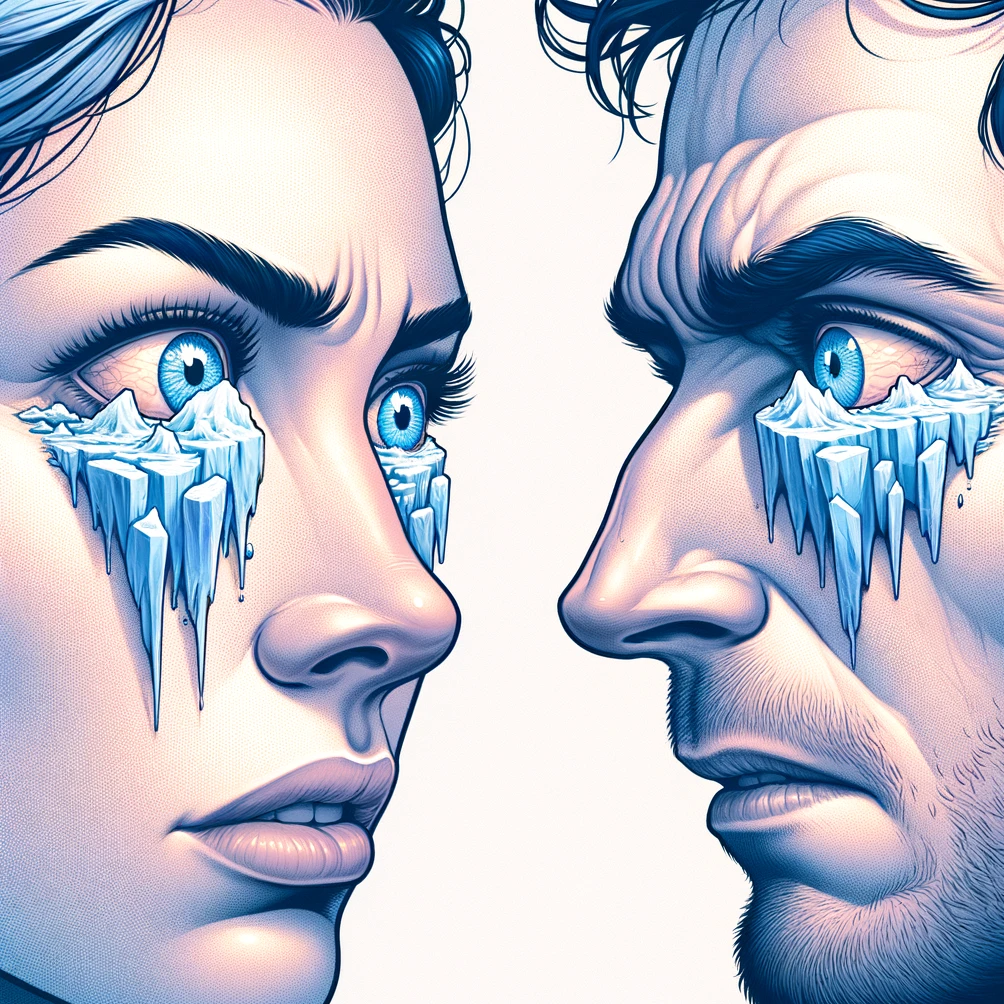}} & 
They looked at each other in an almost hostile way. & 
\textbf{Predicted Label:} Entailment \newline 
\textit{Predicted Explanation:}  \textcolor{red}{The image depicts two characters looking at each other with a tense and confrontational expression}, which is consistent with the claim [...]\\
\midrule
\textbf{Unsound} \newline (\textit{concludes that large amount of money + upward trending chart implies rough waters}) & 
\raisebox{-\totalheight}{\includegraphics[scale=0.06]{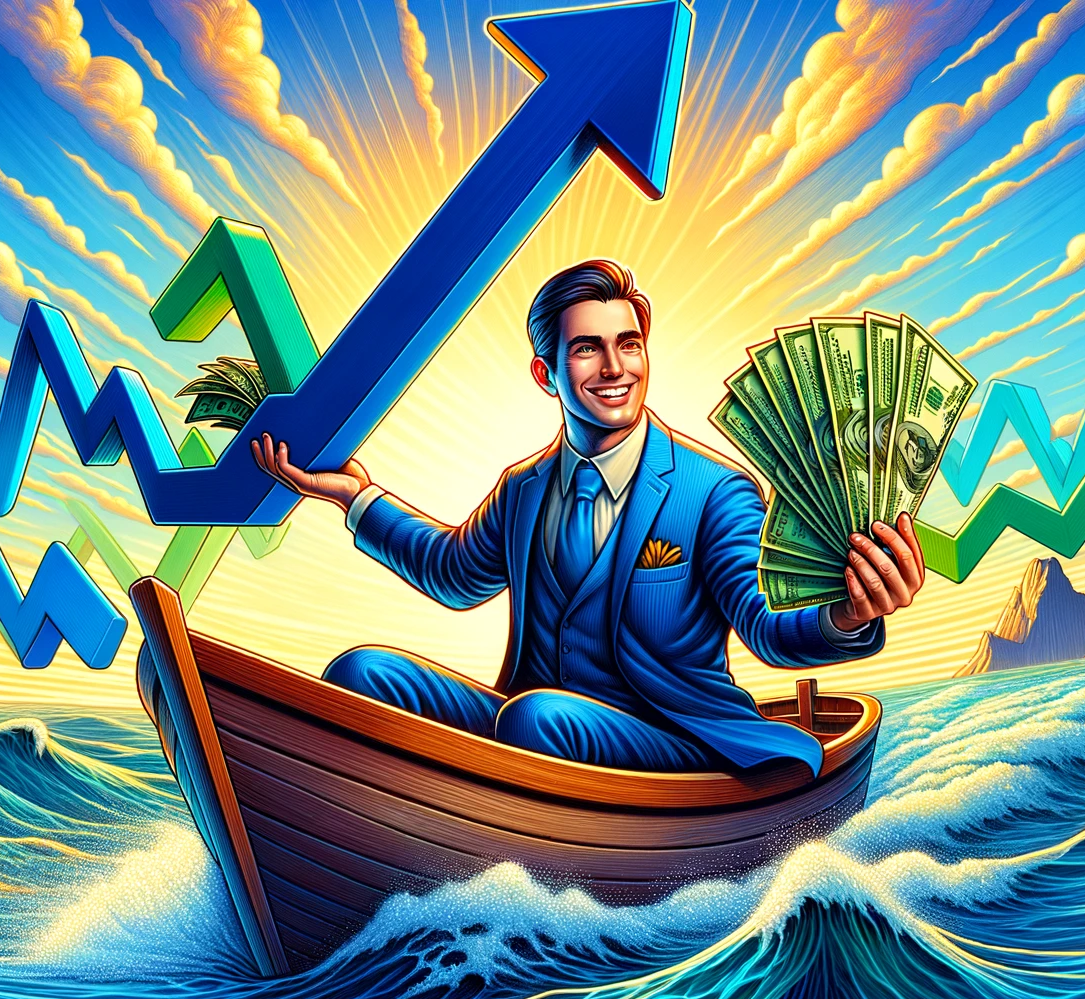}} & 
The world recession of 2001 was one of the worst in history. & 
\textbf{Predicted Label:} \textcolor{red}{Entailment} \newline 
\textit{Predicted Explanation:} The image depicts a man in a boat, holding a large amount of money and a chart with an arrow pointing upwards. \textcolor{red}{This suggests that the man is navigating through rough waters} [...] \\
\bottomrule
\end{tabularx}
\end{adjustbox}
\caption{Examples of error types generated explanations.}
\label{tab:errTypes}
\end{table*}

We conduct human evaluation of generated explanations to assess their quality and identify key errors in reasoning about multimodal figurative meaning. We recruit two expert annotators with background in linguistics for the task and sample 95 random instances from the test set. For each instance, we first provide the annotators with the image, caption and reference explanation and ask the annotators to choose the right label. If the annotator succeeds, they can view the rest of the task, which consists of 3 explanations from our top models by F1@0 in each category: LLaVA-eViL-VF, LLaVA-34B-SG, GPT-4-5shot. The explanations are taken for both correct and incorrect model predictions. For each explanation, we ask whether the explanation is adequate (accurate, correct, complete and concise). 
If not, we ask them to identify one of the errors based on the following taxonomy:

\begin{itemize}[leftmargin=*]
    \itemsep0em 
    \item \textbf{Hallucination:} explanation is not faithful to the image, indicating difficulties with visual comprehension (e.g., generates ``blunt tip'' when the pencil tip is actually sharp in row 1 of Table \ref{tab:errTypes}).
    \item \textbf{Unsound reasoning:} sentences do not adhere to natural logic or violate common sense (e.g., concluding that an upwards arrow and lots of money imply an economic crisis, see row 3).
    \item \textbf{Incomplete reasoning:} while overall the explanation makes sense, it does not address the key property reasons why the image entails or contradicts the caption (for example, does not address the figurative part in the image, see row 2).
    \item \textbf{Verbosity:} the explanation is too verbose.
\end{itemize}

\subsection{How Do Models Perform According to Humans?}

\begin{table}[htbp]
\small
    \centering
    \begin{tabular}{@{}llll@{}}
\toprule
 & \textbf{\begin{tabular}[c]{@{}l@{}}LLaVA-7B\\ eViL+VF\end{tabular}} & \textbf{\begin{tabular}[c]{@{}l@{}}LLaVA-34B\\ SG\end{tabular}} & \textbf{\begin{tabular}[c]{@{}l@{}}GPT-4\\ (5 shot)\end{tabular}} \\ \midrule
\textbf{Adequate \%} & {33.78} & 29.85 & 50.67 \\
\textbf{Preference \%} & {23.08} & 7.69 & 44.23 \\ \bottomrule
\end{tabular}
    \caption{Adequacy and Preference rates  
    for generated explanations.} 
    \label{tab:humanEvalAdeq}
\end{table}

In Table \ref{tab:humanEvalAdeq}, we show adequacy and preference rates for explanations from the 3 systems, where an explanation is deemed adequate or preferred if both annotators agreed it is, and inadequate if both agreed it is not. 
The average IAA using Cohen's $\kappa$ is 0.47, indicating moderate agreement \cite{cohen1960coefficient}. We observe that the teacher GPT-4 model is leading in terms of the adequacy of the explanations and preference rate, as expected from a larger system. Yet still only half of its explanations are considered adequate, confirming that despite good performance on the F1@0 scores, \textit{the models are not yet capable of producing adequate textual explanations in many instances.}
\footnote{Note that during the human-AI collaborative dataset creation 1) the LLM is conditioned on the correct label, 2) its explanation is edited by an expert annotator.}

\subsection{What Errors Do Models Make?}

\begin{figure}
    \centering
    \includegraphics[width=0.9\columnwidth]{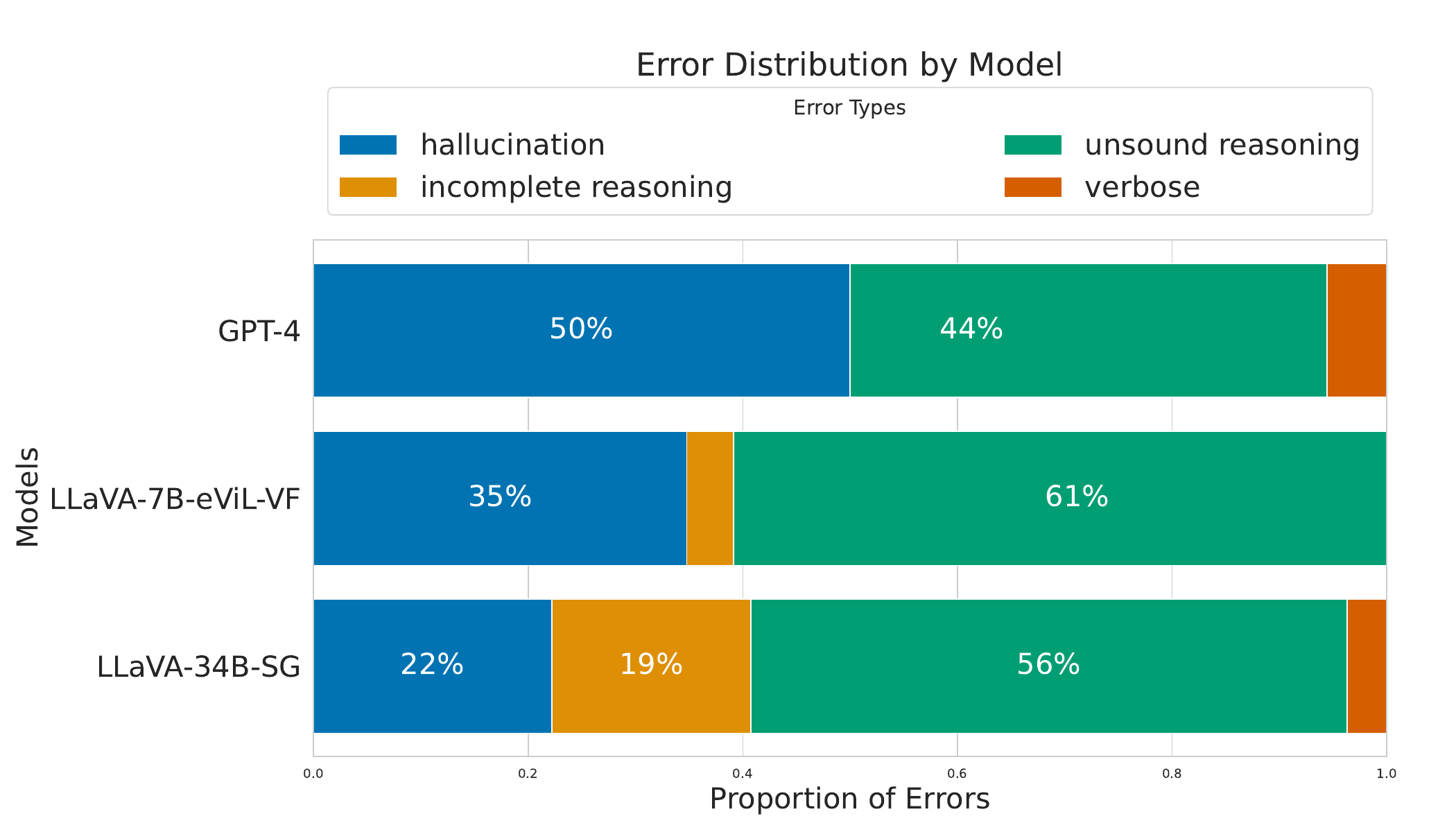}
    \caption{Normalized frequency of main error types in the explanation by model.}
    \label{fig:errorTypesAnnot}
\end{figure}

We perform an analysis of the types of errors from each model when the explanations are considered inadequate in the above evaluation. In Figure \ref{fig:errorTypesAnnot}, we illustrate the normalized frequency of error types when both annotators agree that the explanation is not adequate (i.e., out of all errors for this model, what percentage is each type of error?). Overall, the annotators did not consider verbosity to be a major issue of the systems. For GPT-4, the leading error type is hallucination, indicating the need to improve faithful image recognition even in the most advanced models.
Comparing LLaVA-34B-SG and the fine-tuned model, we see that for the scene graph model a larger percentage of errors is due to incomplete reasoning (possibly due to focusing on the scene graph description rather than the underlying figurative phenomena). For both models, the main error type is unsound reasoning, indicating difficulty for the models to consistently reason about multimodal figurative inputs.

\subsection{How Well Does the Explanation Score Predict Human Judgment on Adequacy?} \label{subsec:explScoreCorr}

We explore whether the proposed explanation score can capture human judgment of explanation adequacy. We collect all instances from Section \ref{sec:humanEval} where both annotators agreed on the adequacy judgement for the explanation.
We evaluate if the explanation score described in Section \ref{subsec:autoMetrics} can act as a good predictor for the human adequacy judgment. We find that the area under the Precision-Recall curve is 0.79, and the maximum F1 score is 0.77, obtainable at the explanation score threshold of 0.53. Hence, we use this threshold to report the results in Table \ref{tab:mainRes}. We also use the threshold of 0.6 since it maximizes F1 such that both precision and recall are above 0.75. 

\subsection{How Well Do Humans Perform?} \label{subsec:humanBaseline}

To find out how humans perform on the task, we hire two expert annotators with formal education in linguistics. We present them with 10 example instances and then ask them to complete 99 randomly sampled test set instances. We also evaluate our best model (see Table \ref{tab:mainRes}) on the same set. Results are shown in Table \ref{tab:humanBaseline}. Human performance is quite strong, almost reaching 90 F1@0 score overall. Human performance is better than our strongest fine-tuned model (LLaVA-7B-eVil+VF) performance with $p < 0.05$ for Annotator 1 and $p<0.07$ for Annotator 2. Humans excel at interpreting memes, with both annotators reaching a 100\% F1 score. Humans also perform noticeably better on the NYCartoons dataset and on the idiom subset of the task. The model has a slight edge in performance on the sarcasm and visual metaphor subsets of the task, perhaps due to difficulty of these subsets and any potential spurious correlations during fine-tuning.

\begin{table}[ht]
\small
    \centering
\begin{tabular}{@{}llll@{}}
\toprule
\textbf{Phenomenon} & \textbf{Dataset} & \textbf{\begin{tabular}[c]{@{}l@{}}Human \\ Avg\end{tabular}} & \textbf{\begin{tabular}[c]{@{}l@{}}LLaVA- \\ eViL+VF\end{tabular}} \\ \midrule
\multirow{2}{*}{\textbf{\begin{tabular}[c]{@{}l@{}}Metaphor\\ /Similes\end{tabular}}} & HAIVMET & 78.84 & \textbf{81.25} \\ \cmidrule(l){2-4} 
 & \begin{tabular}[c]{@{}l@{}}IRFL \\ (metaphor\\ /simile)\end{tabular} & \textbf{94.36} & 77.78 \\ \midrule
\textbf{Idioms} & \begin{tabular}[c]{@{}l@{}}IRFL \\ (idiom)\end{tabular} & \textbf{89.26} & 49.74 \\ \midrule
\textbf{Sarcasm} & MuSE & 68.89 & \textbf{85.42} \\ \midrule
\multirow{2}{*}{\textbf{Humor}} & MemeCap & \textbf{100.0} & 78.03 \\ \cmidrule(l){2-4} 
 & NYCartoons & \textbf{71.43} & 47.83 \\ \midrule
\multicolumn{2}{c}{\textbf{Overall}} & \textbf{89.09} & 77.26 \\ \bottomrule
\end{tabular}
    \caption{Human baseline results (F1@0) by phenomenon and source dataset.}
    \label{tab:humanBaseline}
\end{table}

\section{Conclusion}
We introduce a novel dataset for understanding figurative meaning in multimodal input, V-FLUTE, via an explainable visual entailment task.
Our dataset consists of 6,027 $\{$image, caption, label, explanation$\}$ instances covering diverse phenomena. We find that VLMs struggle to generalize from literal to figurative meaning, particularly in images. When figurative meaning is present in the image rather than text, models perform worse. VLMs benefit from the visual information during training to understand visual figurative meaning. Finally, humans still outperform even powerful VLMs overall. We identify three common error types in VLM reasoning about multimodal figurative phenomena: hallucination and incomplete or unsound reasoning.

\section{Acknowledgments}
We would like to thank the annotators for their work, as well as anonymous reviewers for productive discussion and feedback.
This research is supported in part by the Office of the Director of National Intelligence (ODNI), Intelligence Advanced Research Projects Activity (IARPA), via the HIATUS Program contract \#2022-22072200005. The views and conclusions contained herein are those of the authors and should not be interpreted as necessarily representing the official policies, either expressed or implied, of ODNI, IARPA, or the U.S. Government. The U.S. Government is authorized to reproduce and distribute reprints for governmental purposes notwithstanding any copyright annotation therein.

\section{Ethics}

Following prior work in human-AI collaboration for complex text and image generation \cite{chakrabarty-etal-2022-flute, chakrabarty-etal-2023-spy, ch-wang-etal-2023-sociocultural, saakyan2023iclef}, we opt for an expert-AI collaboration framework where experts edit the initial generations by the language model. Expert feedback is essential to improve the quality of the data, as previous work has identified that crowdworkers on platforms such as Amazon Mechanical Turk could be unreliable for open-ended generation tasks \cite{karpinska-etal-2021-perils}, and might even rely on ChatGPT to provide their answers \cite{veselovsky2023artificial}.  To mitigate these effects, in this work, annotators were recruited through the Upwork platform, allowing to select for relevant level of expertise and verify, e.g., educational and professional background of the annotators. All recruited annotators have significant background in figurative language understanding and have formal educational background in linguistics or literature. All of the annotators are fluent or native/bilingual level in English.  Workers on UpWork were informed that that the work they were doing was going to be used for research purposes. All are fairly compensated with USD \$20 to \$25 per hour with self-reported time needed to complete the tasks. The total budget for the annotation and GPT-4 generations was $\approx \$5,000$ USD. We estimate that it would take approximately 3 times longer to complete the annotation task without the pre-generated explanation, so we estimate that the cost would have at least tripled if the human-AI collaboration approach was not utilized. Workers were paid their wages in full immediately upon the completion of their work. All data collected by human respondents were fully anonymized. We do not report demographic or geographic information, given the limited number of respondents, so as to maintain full anonymity.

\section{Limitations}

We would like to acknowledge the following limitations of our work. 
The textual explanations 
in V-FLUTE dataset were generated with the help of the strongest LLM available at the time of writing the paper, GPT-4. Despite our best efforts in mitigating biases with expert human verification, idiosyncrasies pertaining to GPT-4 outputs may still be present in the text. This means that it is potentially possible for the underlying biases of source datasets of language model generations to propagate into our resource, which we wish to mitigate by carefully examining each dataset instance by one of the 3 expert annotators.

Reference-based evaluation has fundamental flaws, such as not considering all possible explanations, which would be impossible to collect. However, current reference-free metrics for free-text rationales may still have flaws such as bias toward length or the evaluator LLM \cite{stureborg2024large, raina2024llmasajudgerobustinvestigatinguniversal, huang2024limitationsfinetunedjudgemodels, chiang-lee-2023-closer, wei2024systematicevaluationllmasajudgellm}. When evaluating textual explanations against these references, as is the case with any reference-based evaluation, there may also be a preference towards models which output text closer in distribution to the GPT-4 model. Because of that, it is important to utilize the data set in order to compare models other than the teacher model and pay more attention to the F1@0 scores, which represent simple classification scores and do not require the outputs to be similar in distribution. In terms of pure F1 score performance, GPT-4 underperforms the fine-tuned model, and performs very closely with Gemini and Claude that were not used to generate the data, with less than 2\% difference (see column F1@0, Table 4). Although we showed a relatively high predictive power of automatic explanation scores to predict human judgments (see Section \ref{subsec:explScoreCorr}), future work may focus on increasing reliability of reference-based and reference-free textual explanation evaluation methods. 

We also note that the images from the HAIVMet dataset \cite{chakrabarty-etal-2023-spy} are AI generated. However, the majority of the remaining images in V-FLUTE are not AI generated but are naturally occurring or created by humans. However, to mitigate potential biases from AI-generated images, all instances of the data were examined during the expert verification stage, as described in the article.

Label predictions by language models can vary significantly with slight differences in prompt wording \cite{sclarquantifying}, which is why during fine-tuning and inference we utilize over 20+ different templates of instructions (see Table \ref{tab:instructs}). Nevertheless, it is important to consider the models' explanations to better assess their understanding of the phenomena, which we hope to enable with our explainable figurative visual entailment dataset.

\bibliography{custom}
\clearpage
\appendix



\section{Details on Expert Verification} \label{app:annotDetails}
We follow the same procedure for expert verification of all sub-datasets. We recruit 3 expert annotators with background in figurative language and formal educational background in linguistics or literature on Upwork. We first ask to annotate 10 instances by all 3 annotators to ensure they understand the task. We then ensured a high agreement ($\geq 90\%$ pairwise accuracy) between annotators on a subsample of 100 instances of each dataset, and resolved any disagreements through mutual discussion between the annotators and the authors before proceeding. Finally, each annotator proceeds to annotate roughly $\frac{1}{3}$ of the data. 

We provide the annotation interfaces below for HAIVMET (Figure \ref{fig:vismet_interface}), IRFL (Figure \ref{fig:irfl_interface}), MemeCap (Figure \ref{fig:memecap_interface}) and MuSE (Figure \ref{fig:muse_interface}). In addition, instructions were explained in more detail to the annotators via chat on Upwork (for example, the criteria for correctness and conciseness), and any of their doubts and questions were answered.




\section{Dataset Statistics} \label{app:dataStats}

\begin{figure}[htbp]
    \centering
    \includegraphics[width=\columnwidth]{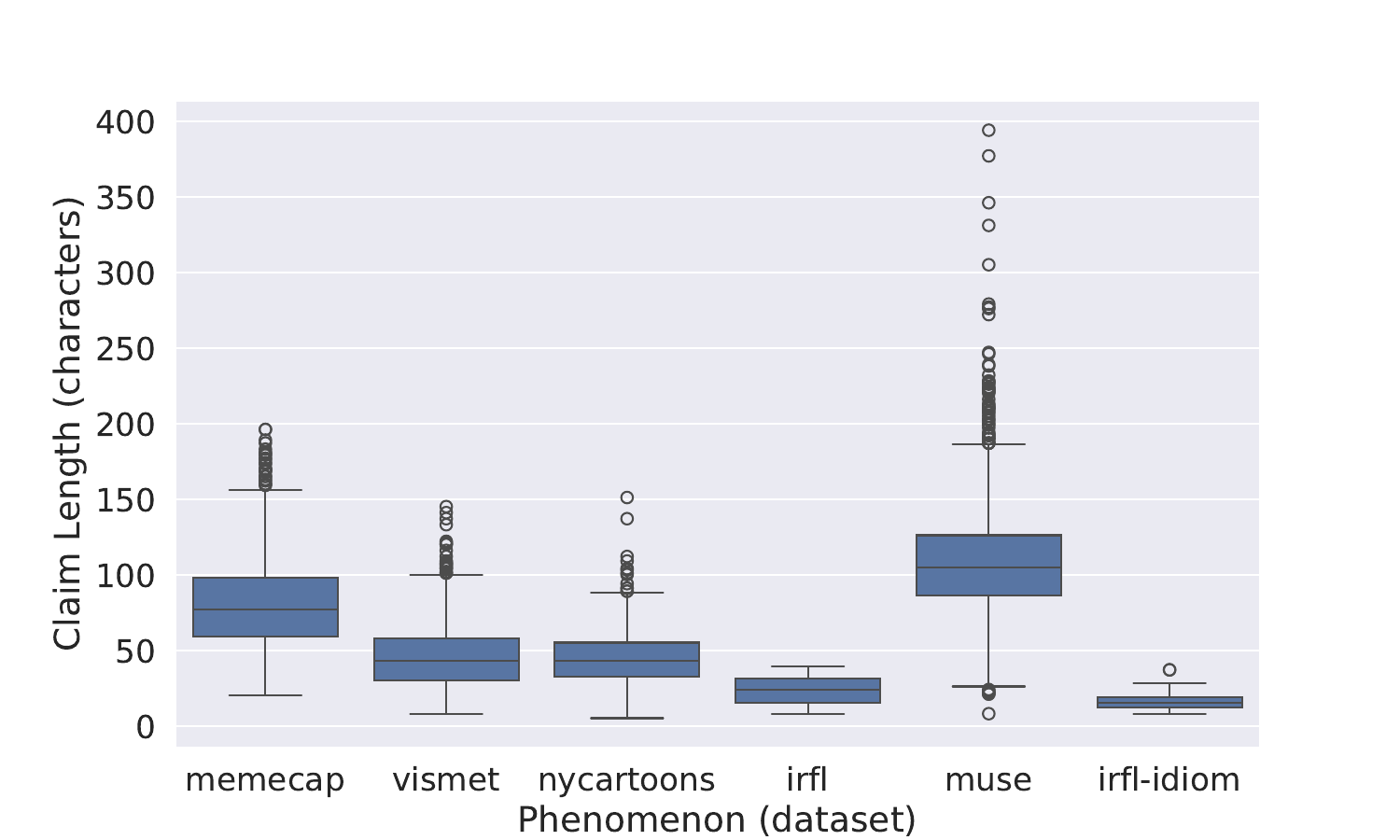}
    \caption{Distribution of lengths of captions by source dataset.}
    \label{fig:claimLen}
\end{figure}

\begin{figure}[htbp]
    \centering
    \includegraphics[width=\columnwidth]{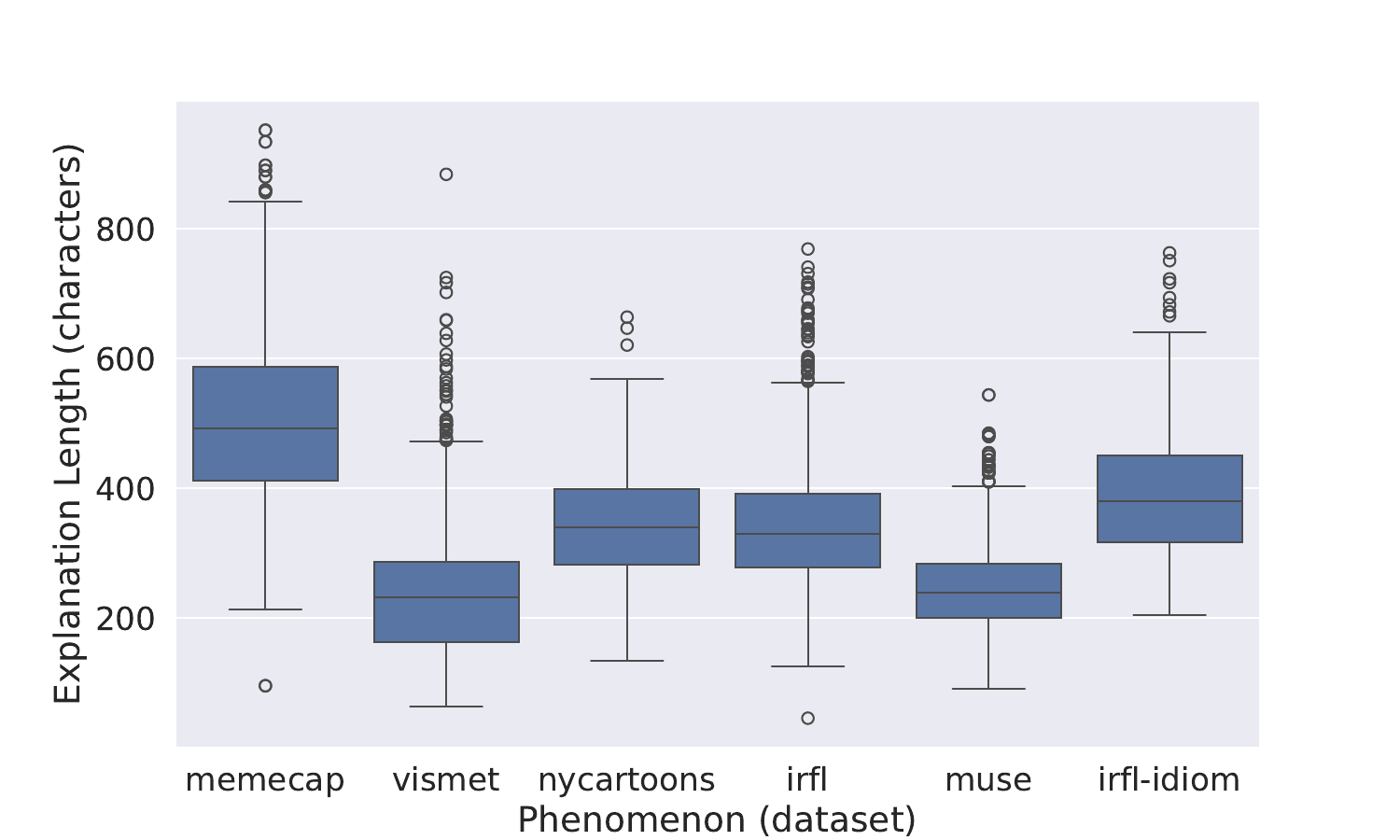}
    \caption{Distribution of lengths of explanations by source dataset.}
    \label{fig:explLen}
\end{figure}

\paragraph{Length distribution}
Average length of a caption in V-FLUTE is $\approx$ 61 characters. Average length of an explanation is $\approx$ 367 characters. Figure \ref{fig:claimLen} shows the distribution of caption lengths, and Figure \ref{fig:explLen} shows the distribution of explanation lengths by source dataset. We manually verified that the outlier instances are correct.

\section{API models Hyperparameters} \label{app:api-hyper}
\subsection{Claude}
\begin{itemize}
    \item Model Name: \texttt{claude-3-opus-20240229}
    \item Max Tokens: 256
    \item Images greater than 5MB were resized maintaining aspect ratio
\end{itemize}

\subsection{GPT-4}
\begin{itemize}
    \item Model Name: \texttt{gpt-4-1106-vision-preview}
    \item Max Tokens: 256
    \item Seed: 42
    \item Image URL detail: 'high'
\end{itemize}

\subsection{Gemini}
\begin{itemize}
    \item Model Name: \texttt{gemini-pro-vision}
    \item Max Tokens: 256
    \item Safety Settings: 'BLOCK NONE'
    \item Images greater than 5MB were resized maintaining aspect ratio
\end{itemize}

\section{Fine-tuning Hyperparameters} \label{app:hyper}

LLava-v1.6-6B and 34B respectively utilize instruction-tuned LLMs as their backbone, Mistral-7Binstruct\footnote{\href{https://huggingface.co/mistralai/Mistral-7B-Instruct-v0.1}{huggingface.co/mistralai/Mistral-7B-Instruct-v0.1}} and Yi-34B\footnote{\href{https://huggingface.co/NousResearch/Nous-Hermes-2-Yi-34B}{huggingface.co/NousResearch/Nous-Hermes-2-Yi-34B}}.

We utilize LoRA \cite{hu2022lora} to fine-tune the models. We utilize the same hyperparameters for all fine-tunes outlined in Appendix \ref{app:hyper} and use early stopping based on a V-FLUTE validation set to prevent overfitting. For evil and e-ViL+V-FLUTE we only fine-tuned for 2 epochs due to size of the e-ViL dataset and took the best checkpoint based on early stopping on V-FLUTE validation set. For eViL we only fine-tuned for 1 epoch to prevent overfitting. For VFLUTE, we trained for 3 epochs, and VFLUTE -Image for 10 epochs (to ensure performance does not increase even with larger number of epochs), for both we took the best checkpoint based on early stopping.

We utilize 4 NVIDIA A100 40GB GPUs for all experiment.

\subsection*{Fine-tuning}
\begin{itemize}
    \item Seed: 42
    \item Vision Tower: openai-clip-vit-large-patch14-336
    \item Number of Training Epochs: 3
    \item Train Batch Size (per device): 16
    \item Eval Batch Size (per device): 4
    \item Learning Rate: 2e-5
    \item Weight Decay: 0
    \item Warmup Ratio: 0.03
    \item Scheduler Type: cosine
    \item Number of epochs: 4 for eViL and eViL + vFLUTE, 10 for VFLUTE
    \item mm-projector-type: mlp2x gelu 
    \item mm-vision-select-layer: -2 
    \item mm-use-im-start-end: False 
    \item mm-use-im-patch-token: False 
    \item image-aspect-ratio: pad 
    \item group-by-modality-length: False
\end{itemize}

\subsection*{LoRA}
\begin{itemize}
    \item lora r: 128 
    \item lora alpha: 256
    \item mm-projector-lr: 2e-5
\end{itemize}

\subsection*{Deepspeed Configuration}
\begin{itemize}
    \item FP16 enabled: auto
    \item BF16 enabled: auto
    \item Micro Batch Size Per GPU: auto
    \item Train Batch Size: auto
    \item Gradient Accumulation Steps: auto
    \item Zero Optimization Stage: 3
\end{itemize}

\subsection*{Training and Inference Instructions}
All models are evaluated using beam search with $n=3$, temperature $0$, max length $256$. In the case of generating scene graphs for the compositional chain-of-thought method, we set the max length to 256 for the graph generation step as recommended by \citet{mitra2023compositional}. API models are evaluated with default hyperparameters.
We format all fine-tuning data in the instruction format following LLaVA \cite{liu2023improvedllava}. To avoid overfitting on a particular instruction for this task, we generate 20 similar instructions using an LLM (ChatGPT-4) and randomly assign one of them to every instance in the training, validation, and testing set. Same instructions were sampled for the e-ViL dataset. Table \ref{tab:instructs} shows the 20 instructions used. 

The instructions were almost always followed. If they were not followed during the data creation process, we discarded those instances. For evaluation, we looked at the sample outputs of each model and designed rules to extract the label and the explanation from the output, which was not too difficult since mostly the instructions were followed well. In the rare cases the model failed to follow instructions, that label would likely be incorrect.

\begin{table*}[htbp]
\centering
\begin{adjustbox}{width=1.8\columnwidth, center}
\begin{tabular}{@{}lp{2\columnwidth}@{}}
\toprule
\textbf{No.} & \textbf{Instruction} \\
\midrule
1 & Does the image's narrative confirm or disprove the claim REPLACE\_CLAIM? Discuss your reasoning and identify it as either entailment or contradiction. \\\addlinespace
2 & Does this image confirm or deny the claim REPLACE\_CLAIM? Discuss your reasoning and determine a label: entailment or contradiction. \\\addlinespace
3 & Is the image's message supporting or opposing the claim REPLACE\_CLAIM? Discuss your rationale and determine the appropriate label: entailment or contradiction. \\\addlinespace
4 & Is there agreement or disagreement between the image and the claim REPLACE\_CLAIM? Provide your analysis and choose between entailment or contradiction. \\\addlinespace
5 & Does the visual evidence support or counter the claim REPLACE\_CLAIM? Provide your explanation and assign it a label of entailment or contradiction. \\\addlinespace
6 & Does the image agree with or dispute the claim REPLACE\_CLAIM? Explain your analysis and mark it as entailment or contradiction. \\\addlinespace
7 & Does the illustration affirm or contest the claim REPLACE\_CLAIM? Provide your argument and choose a label: entailment or contradiction. \\\addlinespace
8 & Is the visual content in agreement or disagreement with the claim REPLACE\_CLAIM? Offer your explanation and categorize it under entailment or contradiction. \\\addlinespace
9 & Is the image in harmony with or in conflict with the statement REPLACE\_CLAIM? Explain your justification and label it as entailment or contradiction. \\\addlinespace
10 & Is the portrayal in the image consistent with or contradictory to the claim REPLACE\_CLAIM? Offer your insights and select between entailment or contradiction. \\\addlinespace
11 & Does the image's depiction validate or refute the claim REPLACE\_CLAIM? Explain your point of view and select a label: entailment or contradiction. \\\addlinespace
12 & Is the content of the image endorsing or challenging the claim REPLACE\_CLAIM? Justify your position and label it as entailment or contradiction. \\\addlinespace
13 & Is the image consistent with the statement REPLACE\_CLAIM? Justify your answer and classify it as either entailment or contradiction. \\\addlinespace
14 & Does the illustration affirm or negate the claim REPLACE\_CLAIM? Articulate your reasoning and apply a label: entailment or contradiction. \\\addlinespace
15 & Does the picture support or refute the assertion REPLACE\_CLAIM? Offer your rationale and select a label: entailment or contradiction. \\\addlinespace
16 & Is the visual portrayal compatible with or adverse to the claim REPLACE\_CLAIM? Justify your viewpoint and label it as entailment or contradiction. \\\addlinespace
17 & Does the image corroborate or dispute the claim REPLACE\_CLAIM? Outline your reasoning and categorize it under entailment or contradiction. \\\addlinespace
18 & Is the depiction aligned with or against the claim REPLACE\_CLAIM? Share your evaluation and identify it as either entailment or contradiction. \\\addlinespace
19 & Does the image entail or contradict the claim REPLACE\_CLAIM? Explain your reasoning and provide a label between entailment or contradiction. \\\addlinespace
20 & Can the image be seen as validating or opposing the claim REPLACE\_CLAIM? Explain your thought process and assign a label of entailment or contradiction \\\addlinespace
21 & Is the image’s representation supportive of or contradictory to the claim REPLACE\_CLAIM? Articulate your analysis and assign the label: entailment or contradiction. \\\addlinespace
\bottomrule
\end{tabular}
\end{adjustbox}
\caption{Instruction variations for the figurative visual entailment task.}
\label{tab:instructs}
\end{table*}

\subsection*{Evaluation Hyperparameters}
Following prior work, we utilize BERTScore \cite{bertscore} 
 based on the \texttt{microsoft-deberta-xlarge-mnli} model \cite{he2021deberta, williams-etal-2018-broad} 
 and BLEURT \cite{sellam-etal-2020-bleurt} 
 based on BLEURT-20 \cite{pu2021learning} for the ExplanationScore.

\section{Prompts for LLMs} \label{app:prompts}

\subsection{HAIVMET}
\subsubsection{One-shot Prompt for generating explanations}
\label{subsubsec:vismet-gen-explain}
We describe our one-shot prompts given to an LLM (\texttt{gpt-3.5-turbo-instruct-0914}) for generating explanations of entailment-contradiction relationship. Refer to Table \ref{tab:haivmet-prompt} for the detailed prompt.
  
\begin{table}[htbp]
\small
\centering
\def\arraystretch{1.05}
\begin{tabular}{|l|}
\hline
\begin{tabular}[c]{@{}l@{}}You will be provided a Caption describing what\\ is in the image in detail. You will also be\\ provided with a Claim that contradicts or\\ is entailed by the image (as indicated by\\ the Label). Your task is to explain why\\ the claim contradicts or is entailed by\\ the image. Be very brief in your explanation.\\ Start your explanation by describing what the\\ image depicts, displays or shows.\end{tabular} \\ \hline
\begin{tabular}[c]{@{}l@{}}Caption: An illustration of a group of soldiers\\ with red skin, horns, and pitchforks in hand\\ with a fierce expression on their faces. \\
Claim: The soldiers were angels.\\
Label: Contradiction \\
Explanation: {\color{blue}The image depicts soldiers } \\ {\color{blue}with red skin, horns, and pitchforks, which} \\ {\color{blue}are traditional characteristics associated } \\ {\color{blue}with demons, not angels. Therefore, the} \\ {\color{blue}claim that the soldiers were angels contradicts } \\ {\color{blue}the image.}\end{tabular} \\ \hline
\begin{tabular}[c]{@{}l@{}}Caption:......\end{tabular} \\ \hline
\end{tabular}
\caption{ One shot prompt given to an LLM (\texttt{gpt-3.5-turbo-instruct-0914}) for generating explanations of entailment-contradiction relationship of the HAIVMET dataset.}
\label{tab:haivmet-prompt}
\end{table}

\subsection{IRFL}

\subsubsection{Zero-shot Prompt for generating explanations}

We provide our zero-shot prompt given to an LLM (\texttt{gpt-4-vision-preview}) for generating the entailment explanations given the claim and the image. Refer Table \ref{tab:irfl-prompt-gen-explain} for the detailed prompt.

\begin{table}[htbp]
\small
\centering
\def\arraystretch{1.05}
\begin{tabular}{|l|}
\hline
\begin{tabular}[c]{@{}l@{}}You will be provided an image. You will also be \\provided with a {\color{blue}simile} that contradicts or is \\ entailed by the image (as indicated by the Label).\\Your task is to explain why the {\color{blue}simile} contradicts \\or is entailed by the image. Be very brief in your \\explanation and remain consistent to the Label in \\your explanation. Start your explanation by \\describing what the image depicts, displays or \\
shows. \\
{\color{blue}Simile}: .... \\
Label: .... \\
Explanation: \end{tabular} \\ \hline
\end{tabular}
\caption{ Zero shot prompt given to an LLM (\texttt{gpt-4-vision-preview}) for generating explanations of entailment-contradiction relationship of the IRFL Dataset. The dataset contains similes, metaphors and idioms. For metaphors and idioms, the word simile in the prompt is replaced with the corresponding type.}
\label{tab:irfl-prompt-gen-explain}
\end{table}

\subsection{MuSE}
\label{sub:muse-prompts}
\subsubsection{Few-shot Prompt for generating opposite claims}
We provide our few-shot prompt given to an LLM (\texttt{(\texttt{gpt-4-0613})}) for generating the opposite claims. Refer Table \ref{tab:muse-gen-opp} for the detailed prompt.

\begin{table}[htbp]
\small
\centering
\def\arraystretch{1.05}
\begin{tabular}{|l|}
\hline
\begin{tabular}[c]{@{}l@{}}u are an online redditor or flickr user and u\\ always type in informal style. Convert the \\following sarcastic claim into a non-sarcastic\\ claim. Preserve the informal style, including\\ capitalization. Be super laid back and informal!!!
\end{tabular} \\ \hline

\begin{tabular}[c]{@{}l@{}}1. Sarcastic claim: stairs vs . escalator in 
 \\airport . i wonder why we have an \# obesity \\ problem ? \# publichealth \# ncds \# globalhealth \\ \# isometimesdothistoo\\ 
 Explanation: no wonder we have an obesity \\ problem since everyones using escalator \\ instead of stairs in airport.\\
Non-sarcastic claim: {\color{blue}it s clear why we have an  } \\ {\color{blue} \# obesity problem look at stairs vs. escalator   } \\ {\color{blue}in airport }\end{tabular} \\ \hline
\begin{tabular}[c]{@{}l@{}}Claim:......\end{tabular} \\ \hline
\begin{tabular}[c]{@{}l@{}}Explanation:......\end{tabular} \\ \hline
\end{tabular}
\caption{ Few shot prompt given to an LLM (\texttt{gpt-4-0613}) for generating opposite claims utilizing the sarcastic claim and crowd worker explanation.}
\label{tab:muse-gen-opp}
\end{table}

\subsubsection{Zero-shot Prompt for Rephrasing}

We provide our zero-shot prompt given to an LLM (\texttt{gpt-4-vision-preview}) for rephrasing the explanations given the claim and the crowd worker explanation. Refer Table \ref{tab:muse-gen-explain} for the detailed prompt.

\begin{table}[htbp]
\small
\centering
\def\arraystretch{1.05}
\begin{tabular}{|l|}
\hline
\begin{tabular}[c]{@{}l@{}}Paraphrase the draft explanation of why the image\\ contradicts the literal interpretation of the claim.\\ Be sure to first describe the image in one sentence.\\ Keep your answer short. Do not refer to the claim\\ or the draft explanation in your paraphrase. Stay \\ close to the draft explanation. \\
Claim: .... \\
Draft Explanation: \end{tabular} \\ \hline
\end{tabular}
\caption{ Zero shot prompt given to an LLM (\texttt{gpt-4-vision-preview}) for rephrasing the explanations given the claim and the.}
\label{tab:muse-gen-explain}
\end{table}

\subsection{MemeCap} \label{subsec:memecap-prompts}

\subsubsection{Few-shot Prompt for generating entailing claims} \label{subsubsec:gen-prompt-ent}

We describe our few-shot prompts given to an LLM (\texttt{gpt-4-0613}) for generating entailing captions as part of the pipeline. Refer to Table \ref{tab:memecap-gen-claims} for the detailed prompt.
  
\begin{table}[htbp]
\small
\centering
\def\arraystretch{1.05}
\begin{tabular}{|l|}
\hline
\begin{tabular}[c]{@{}l@{}}You will be provided with a meme caption. Your\\ task is to write the meme caption as a claim such\\that the meme poster is not mentioned in the\\ claim.\end{tabular} \\ \hline
\begin{tabular}[c]{@{}l@{}}Caption: Meme poster is saying that searching \\Google plus the term you want to search on \\reddit is better than searching reddit itself.\\ Claim: {\color{blue}Searching on Google with the term } \\ {\color{blue} you want to search plus 'reddit' is more effective } \\ {\color{blue} than searching directly on Reddit. }\end{tabular} \\ \hline
\begin{tabular}[c]{@{}l@{}}Caption: The person who wrote the post is saying\\ people on Instagram are soft and reddit are funny.\\ Claim: {\color{blue}People on Instagram are soft, whereas  } \\ {\color{blue}those on Reddit are funny.} \end{tabular} \\ \hline
\begin{tabular}[c]{@{}l@{}}Caption:......\end{tabular} \\ \hline
\end{tabular}
\caption{ Two shot prompt given to an LLM (\texttt{gpt-4-0613}) for generating entailing claims utilizing the meme captions part of the MemeCap dataset.}
\label{tab:memecap-gen-claims}
\end{table}

\subsubsection{Zero-shot Prompt for validating the entailing captions}

We describe our zero-shot prompt given to an LLM (\texttt{gpt-4-vision-preview}) for validating the claims generated in the previous step.  Refer Table \ref{tab:memecap-validate-claims} for the detailed prompt.

\begin{table}[htbp]
\small
\centering
\def\arraystretch{1.05}
\begin{tabular}{|l|}
\hline
\begin{tabular}[c]{@{}l@{}}You will be provided a meme image and a claim. \\Your task is to check whether the claim entails the \\image. Answer with a Yes or No. \\
Claim: .....\end{tabular} \\ \hline
\end{tabular}
\caption{ Zero shot prompt given to an LLM (\texttt{gpt-4-vision-preview}) for validating the claims generated in \ref{subsubsec:gen-prompt-ent}. The corresponding meme image is also attached with the prompt.}
\label{tab:memecap-validate-claims}
\end{table}

\subsubsection{Few-shot Prompt for generating opposite claims}
We provide our few-shot prompt given to an LLM (\texttt{(\texttt{gpt-4-0613})}) for generating the opposite claims. Refer Table \ref{tab:memecap-gen-opp} for the detailed prompt.

\begin{table}[htbp]
\small
\centering
\def\arraystretch{1.05}
\begin{tabular}{|l|}
\hline
\begin{tabular}[c]{@{}l@{}}Claim: A useful feature has been removed \\ on YouTube, causing disappointment. \\Explanation: The image shows a painting\\ of a character with a distraught face and\\ a speech bubble that reads "y tho," placed\\ over text saying "When YouTube removed\\ sort by oldest option." This implies that the\\ removal of the sort by oldest option is a\\ decision that users are questioning, hence\\ indicating disappointment over the loss of\\ a useful feature.\\ Opposite claim:  {\color{blue}An unhelpful feature has} \\ {\color{blue}  been removed on YouTube, causing happiness.   } \end{tabular} \\ \hline
\begin{tabular}[c]{@{}l@{}}Claim:......\end{tabular} \\ 
\begin{tabular}[c]{@{}l@{}}Explanation:......\end{tabular} \\ \hline
\end{tabular}
\caption{ Few shot prompt given to an LLM (\texttt{gpt-4-0613}) for generating opposite claims utilizing the generated claim and explanation.}
\label{tab:memecap-gen-opp}
\end{table}

\subsubsection{Zero-shot Prompt for generating explanations}

We provide our zero-shot prompt given to an LLM (\texttt{gpt-4-vision-preview}) for generating the entailment explanations given the claim and the image. Refer Table \ref{tab:memecap-gen-explain} for the detailed prompt.

\begin{table}[htbp]
\small
\centering
\def\arraystretch{1.05}
\begin{tabular}{|l|}
\hline
\begin{tabular}[c]{@{}l@{}}You will be provided a meme. You will also be \\ provided with a claim that entails the image. \\Your task is to explain why the claim is entailed \\by the image. Be very brief in your explanation \\ and start your explanation by describing what \\ the image depicts, displays or shows. \\
Claim: .... \\
Explanation: \end{tabular} \\ \hline
\end{tabular}
\caption{ Zero shot prompt given to an LLM (\texttt{gpt-4-vision-preview}) for generating the entailment explanations. The corresponding meme image is also attached with the prompt.}
\label{tab:memecap-gen-explain}
\end{table}

\section{Model Taxonomy}

The taxonomy of all models used for automatic evaluation is shown in Figure \ref{fig:modelTaxonomy}.
\begin{figure*}
\small
    \centering
    \begin{forest}
for tree={
  grow=east,
  draw,
  rectangle,
  align=center,
  parent anchor=east,
  child anchor=west,
  edge={thick},
  l sep+=10pt,
  tier/.pgfmath=level(),
  edge path={
    \noexpand\path [draw, \forestoption{edge}] (!u.parent anchor) -- +(5pt,0) |- (.child anchor)\forestoption{edge label};
  }
}
[Models
    [Off-the-shelf
        [API-based
            [Claude Opus]
            [GPT-4]
            [Gemini]
        ]
        [Open (LLaVA-ZS)
            [7B]
            [7B-SG]
            [34B]
            [34B-SG]
        ]
    ]
    [Fine-tuned (LLaVA-7B)
        [eViL]
        [VFLUTE]
        [eViL+VFLUTE]
    ]
]
\end{forest}
    \caption{Taxonomy of models used for the study.}
    \label{fig:modelTaxonomy}
\end{figure*}

\section{Multimodal Structured Chain-of-Thought Performance} \label{app:CoT}

In addition to zero-shot testing, we also test these models using \textit{Compositional Chain-of-Thought Prompting} proposed by \citet{mitra2023compositional}. The method prompts the model \textit{zero-shot} to generate a scene graph in JSON format and then utilizes that scene graph in another prompt to answer the relevant question. 
We refer to these models as LLaVA-ZS-7B-SG and LLaVA-ZS-34B-SG for the 7B and 34B LLaVA configurations described above. 

\paragraph{Scene graph prompting and few-shot prompting improves performance on the figurative visual entailment task.} Observing the results in Table \ref{tab:mainRes}, we can see that the multimodal few-shot prompting and scene graph prompting, having demonstrated their effectiveness for literal inputs, also show improved performance on the figurative visual entailment task. However, the explanations generated by SG-models tend to overly focus on the contents of the scene graph rather than the underlying figurative phenomena, possibly causing a decrease in explanation score.

\section{Additional Models} \label{app:addlModels}

In addition to the LLaVA architecture, we conduct experiments with the Instruct-BLIP model \cite{10.5555/3666122.3668264}, specifically, the Instruct-BLIP-Vicuna-7B version. As can be seen in Table \ref{tab:blip}, InstructBLIP shows a weaker performance compared to LLaVA-7B, especially in explanation quality (4.14 F1@53 for InstructBLIP while 35.56 for LLaVA-7B-ZS, and 2.07 F1@60 while 18.38 for LLaVA as can be seen in Table \ref{tab:mainRes}). It struggled to generate scene graph descriptions, unlike LLaVA-7B.
Despite extensive instruction-tuning, it performed below a random baseline in our figurative entailment task (F1@0: 43.37).

\begin{table}[ht]
    \centering
    \begin{tabular}{lccc}
        \toprule
        \textbf{Model Name}     & \textbf{f1@0} & \textbf{f1@53} & \textbf{f1@60} \\
        \midrule
        InstructBlip-7B-ZS      & 43.37         & 4.14           & 2.07           \\
        InstructBlip-7B-SG      & 38.03         & 4.15           & 1.38           \\
        \bottomrule
    \end{tabular}
    \caption{F1 Scores for Different Models}
    \label{tab:blip}
\end{table}

We also experimented with a state-of-the-art multimodal model GPT-4o that was released after our dataset was created. As expected, the results are better than those of GPT-4 due to improvements in the multimodal processing of GPT-4o. However, the F1@53 and F1@60 scores suggest there could still be improvements in explanation quality. Compared to the 7B fine-tuned LLaVA model, the zero-shot GPT-4o still underperforms the fine-tuned models in terms of F1@53 and is comparable in terms of F1@0. GPT-4o in the few-shot scenario (5 example) shows better results than the fine-tune model. These results can add to the discussion in our field between smaller open-source models and bigger and proprietary models in terms of performance accuracy and capabilities.

\begin{table}[ht]
    \centering
    \begin{tabular}{lccc}
        \toprule
        \textbf{Model Name} & \textbf{f1@0} & \textbf{f1@53} & \textbf{f1@60} \\
        \midrule
        GPT-4o        & 75.41 & 60.97 & 37.20 \\
        \myquad $\dashrightarrow$ 5-shot & 79.42 & 69.35 & 56.31 \\
        \bottomrule
    \end{tabular}
    \caption{F1 Scores for GPT-4 Models}
    \label{tab:gpt4o}
\end{table}

\section{By-Phenomenon Performance} \label{app:byPhen}

\begin{figure}[ht]
    \centering
    
    \begin{subfigure}[b]{\columnwidth}
        \centering
        \includegraphics[width=\textwidth]{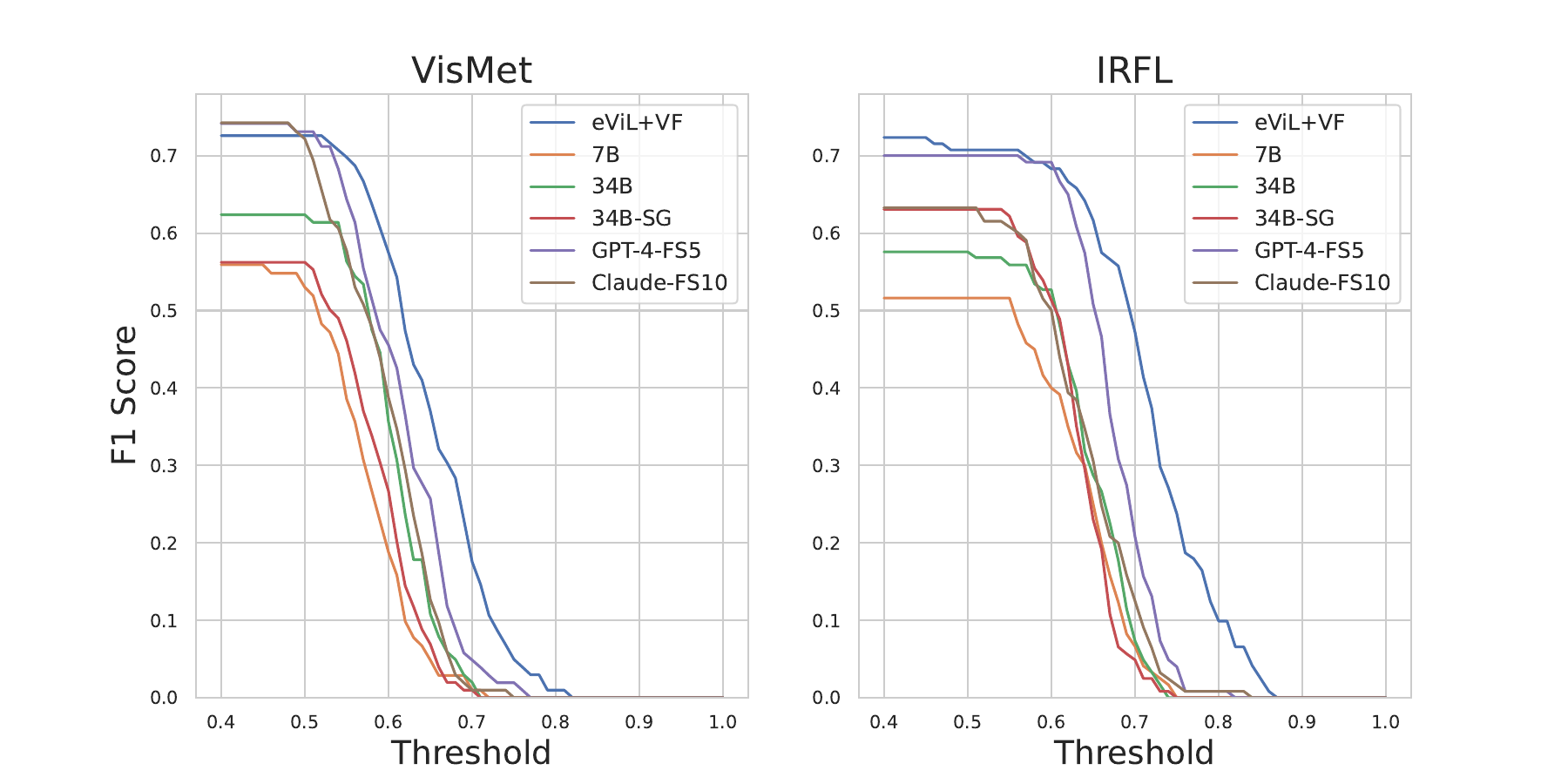}
        \caption{Metaphors and Similes}
        \label{fig:metaphors_similes}
    \end{subfigure}
    \hfill 
    \begin{subfigure}[b]{\columnwidth}
        \centering
        \includegraphics[width=\textwidth]{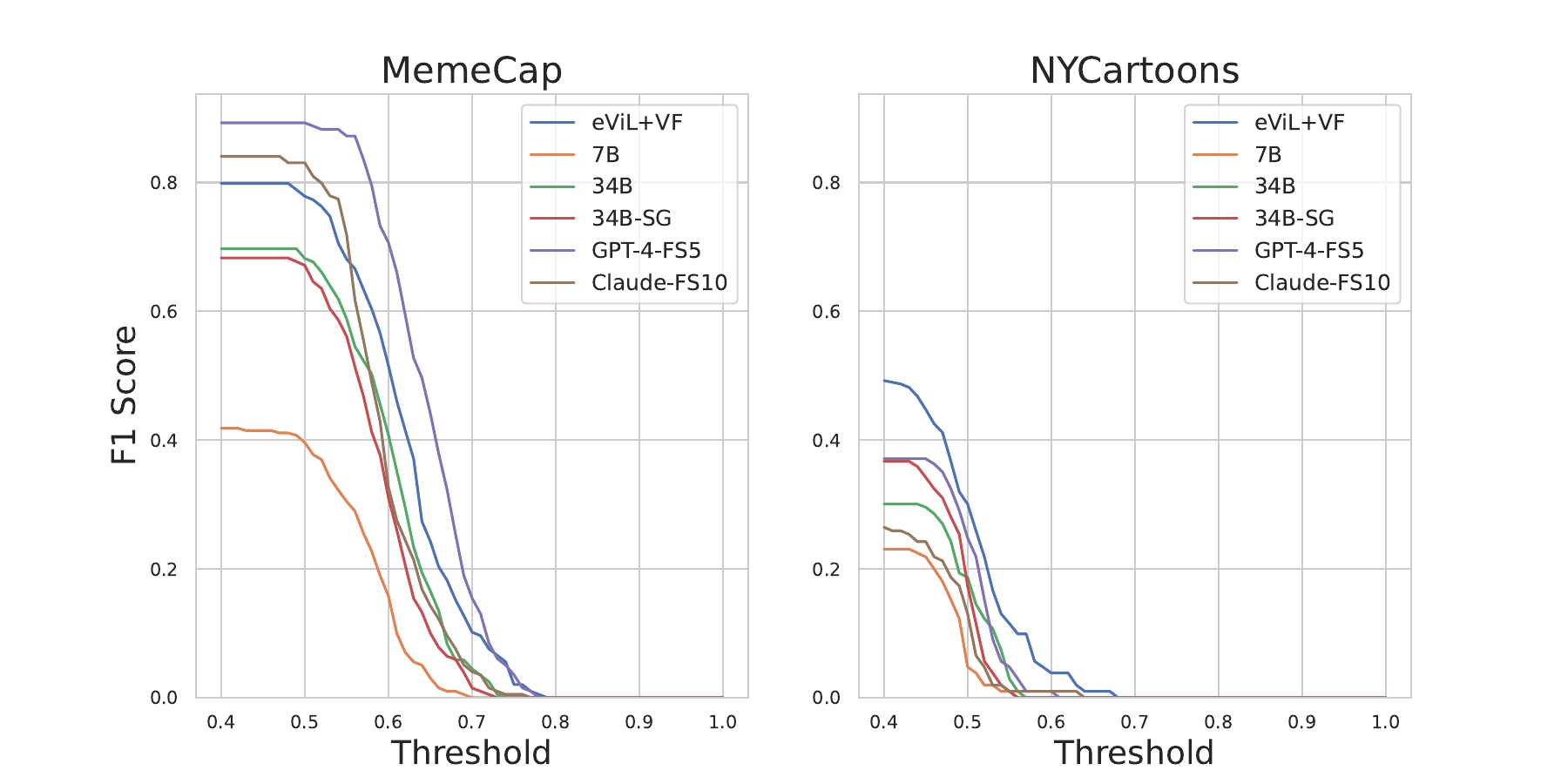}
        \caption{Humor}
        \label{fig:humor}
    \end{subfigure}
    \hfill 
    \begin{subfigure}[b]{\columnwidth}
        \centering
        \includegraphics[width=\textwidth]{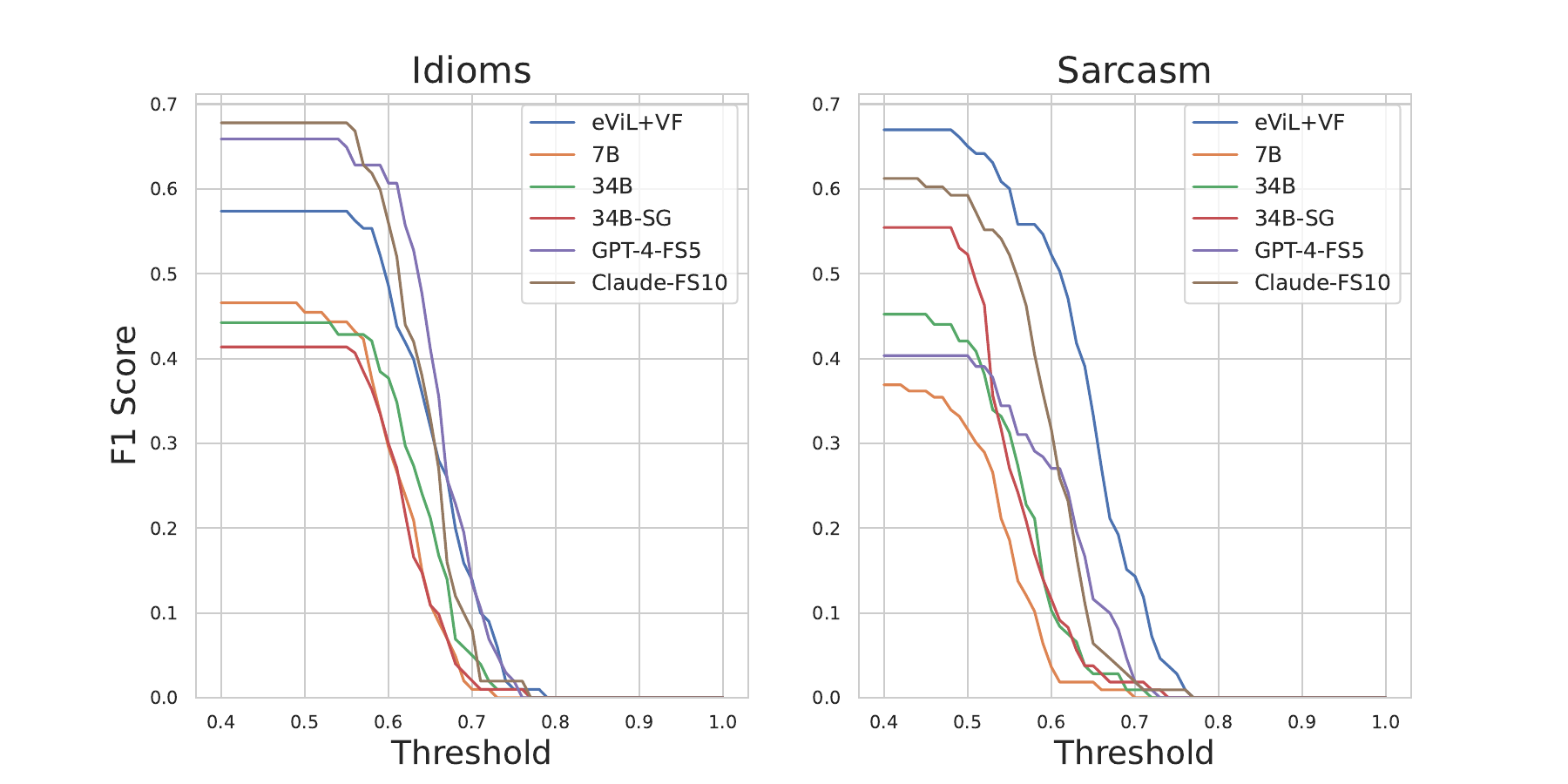}
        \caption{Sarcasm and Idioms}
        \label{fig:sarcasm_idioms}
    \end{subfigure}

    \caption{Performance of the models by phenomenon.}
    \label{fig:byPhen}
\end{figure}

In Figure \ref{fig:byPhen}, we show the performance of the models by phenomenon and dataset across various thresholds.

\section{How Do Models Perform When Only Predicting the Label?}
In our experiments, we found that predicting only the label improves accuracy compared to predicting label and explanation (this is expected and observed in other work on textual explanations such as e-SNLI \cite{eSNLI}). However, these predictions are less reliable since they could be due to spurious correlations (which is why we require the model to generate textual explanations). We also found when fine-tuning the model in a multi-task fashion with explanations (i.e., two tasks, one of generating explanations and one of predicting the label), the accuracy improves compared to when fine-tuning only for the prediction task (F1 score of 80.85 vs. 83.26, $p < 0.1$), in line with previous findings by \citet{hsieh-etal-2023-distilling}.

\section{Annotation Interfaces}

\begin{figure*}[ht]
\centering
    \includegraphics[width=0.9\textwidth]{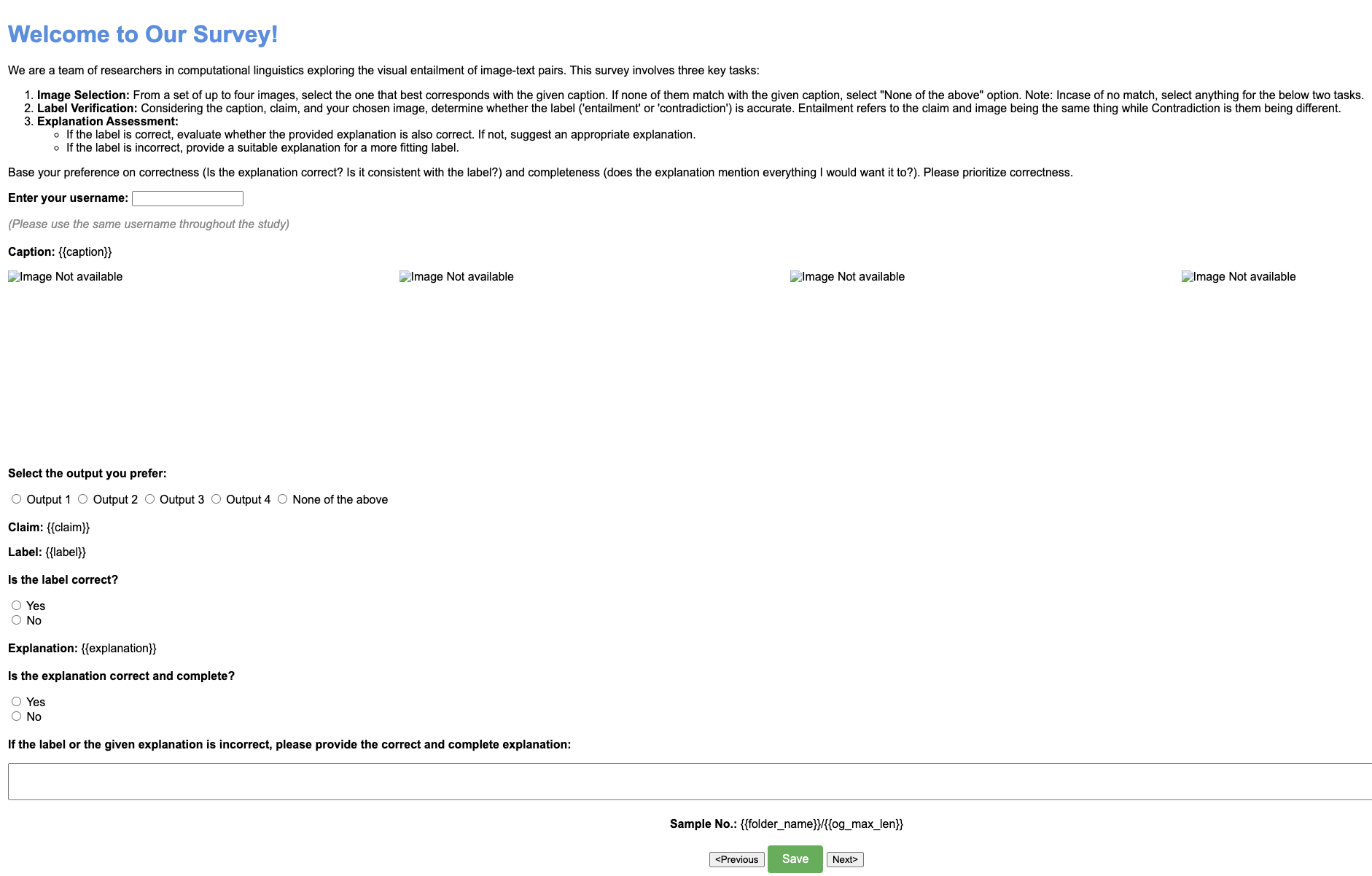}
    \caption{Annotation interface for HAIVMET.}
    \label{fig:vismet_interface}
\end{figure*}

\begin{figure*}[ht]
\centering
    \includegraphics[width=0.9\textwidth]{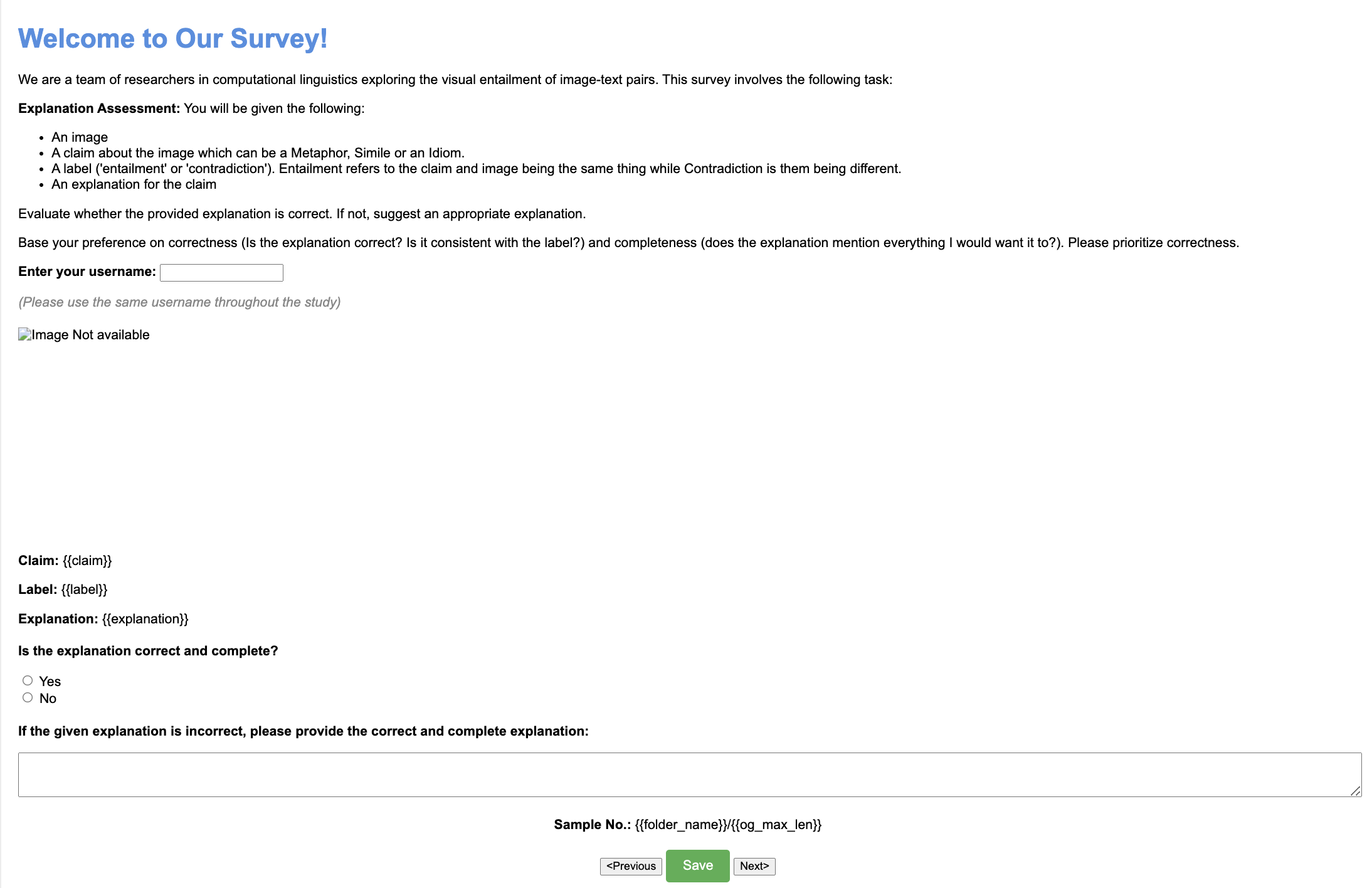}
    \caption{Annotation interface for IRFL.}
    \label{fig:irfl_interface}
\end{figure*}

\begin{figure*}[ht]
\centering
    \includegraphics[width=0.9\textwidth]{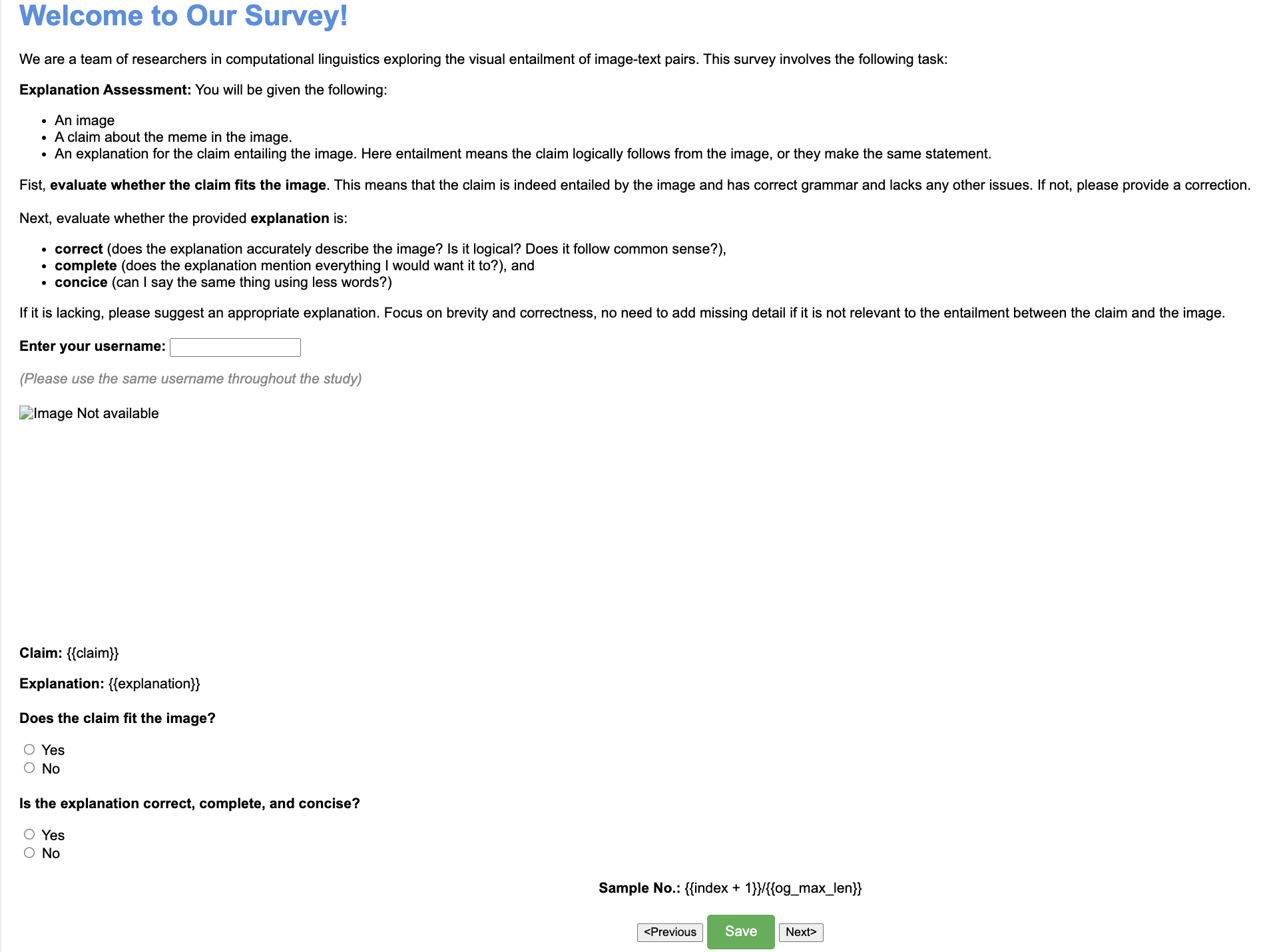}
    \caption{Annotation interface for MemeCap.}
    \label{fig:memecap_interface}
\end{figure*}

\begin{figure*}[ht]
\centering
    \includegraphics[width=0.9\textwidth]{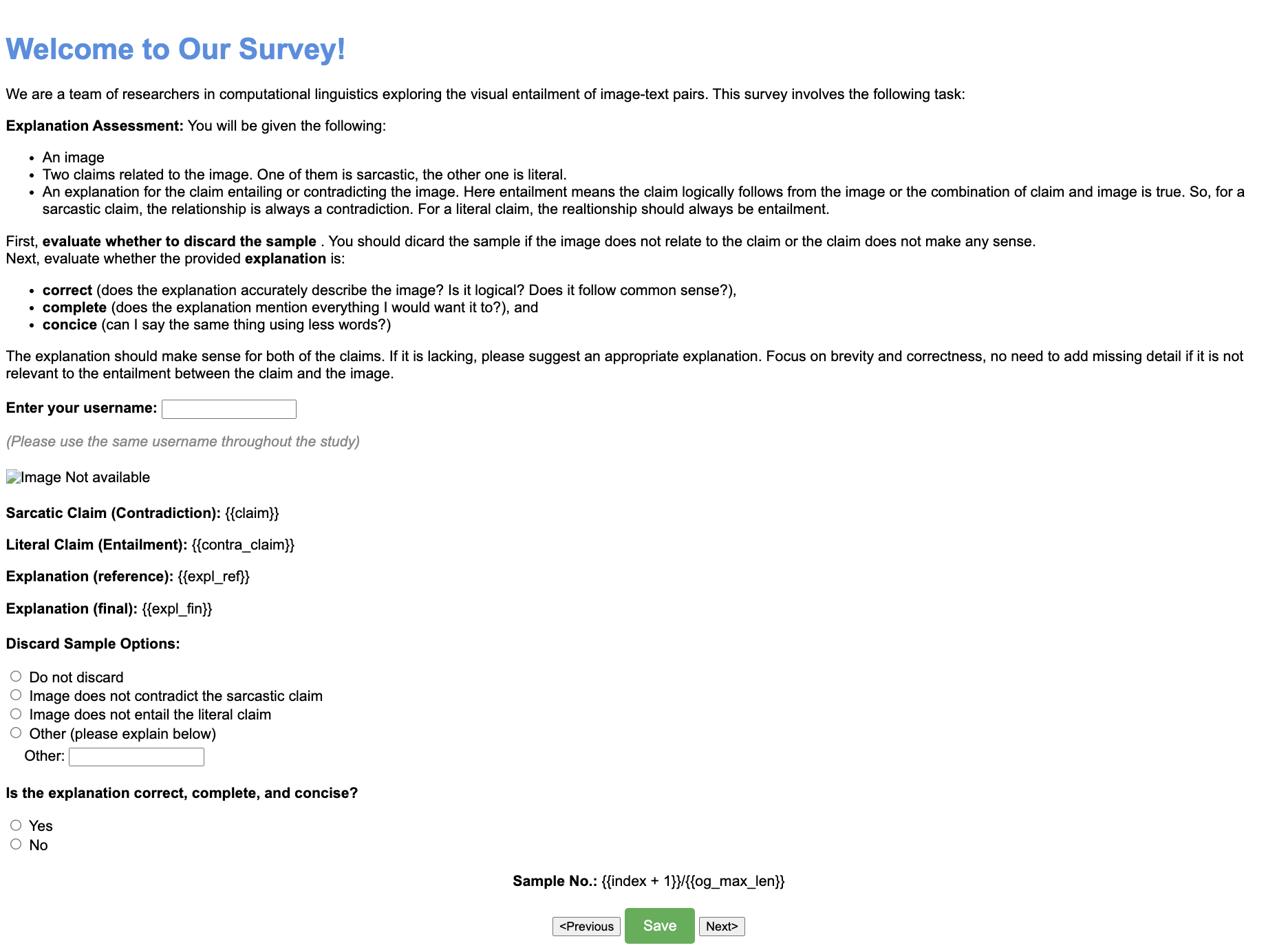}
    \caption{Annotation interface for MuSE.}
    \label{fig:muse_interface}
\end{figure*}

We provide the annotation interfaces below for HAIVMET (Figure \ref{fig:vismet_interface}), IRFL (Figure \ref{fig:irfl_interface}), MemeCap (Figure \ref{fig:memecap_interface}) and MuSE (Figure \ref{fig:muse_interface}). In addition, instructions were explained in more detail to the annotators via chat on Upwork, and any of their doubts and questions were answered.

\end{document}